\documentclass{article}

\usepackage{PRIMEarxiv}

\usepackage[utf8]{inputenc} 
\usepackage[T1]{fontenc}    
\usepackage{hyperref}       
\usepackage{url}            
\usepackage{booktabs}       
\usepackage{amsfonts}       
\usepackage{nicefrac}       
\usepackage{microtype}      
\usepackage{lipsum}
\usepackage{fancyhdr}       
\usepackage{graphicx}       
\graphicspath{{media/}}     

\usepackage{subcaption}
\usepackage{comment}
\title{The Good, the Bad, and the Hulk-like GPT: \\Analyzing Emotional Decisions of Large Language Models in Cooperation and Bargaining Games}

\author{
  Mikhail Mozikov \\
  AIRI, NUST MISIS \\
  Moscow\\
  Russia\\
  \And
  Nikita Severin \\
  HSE University \\
  Moscow\\
  Russia\\
  \And
  Valeria Bodishtianu \\
  Cornell University \\
  USA\\
  \And
  Maria Glushanina \\
  SPbU \\
  Saint-Petersburg\\
  Russia\\
  \And
  Mikhail Baklashkin \\
  independent researcher \\
  Moscow\\
  Russia\\
  \And
  Andrey V. Savchenko \\
  HSE University \\
  Moscow\\
  Russia\\
  \And
  Ilya Makarov \\
  AIRI \\
  Moscow\\
  Russia\\
   \texttt{iam.dadii.dh@gmail.com} \\
}

\begin{document}
\maketitle

\begin{abstract}
Behavior study experiments are an important part of society modeling and understanding human interactions. In practice, many behavioral experiments encounter challenges related to internal and external validity, reproducibility, and social bias due to the complexity of social interactions and cooperation in human user studies. Recent advances in Large Language Models (LLMs) have provided researchers with a new promising tool for the simulation of human behavior. However, existing LLM-based simulations operate under the unproven hypothesis that LLM agents behave similarly to humans as well as ignore a crucial factor in human decision-making: emotions.

In this paper, we introduce a novel methodology and the framework to study both, the decision-making of LLMs and their alignment with human behavior under emotional states. Experiments with GPT-3.5 and GPT-4 on four games from two different classes of behavioral game theory showed that emotions profoundly impact the performance of LLMs, leading to the development of more optimal strategies. While there is a strong alignment between the behavioral responses of GPT-3.5 and human participants, particularly evident in bargaining games, GPT-4 exhibits consistent behavior, ignoring induced emotions for rationality decisions. Surprisingly, emotional prompting, particularly with `anger' emotion, can disrupt the ``superhuman'' alignment of GPT-4, resembling human emotional responses.
\end{abstract}


\section{Introduction}

The last few decades have seen an increasing interest in the field of behavioral economics, with a large body of literature devoted to the idea of enhancing \textit{homo economicus} with psychological factors such as cognitive biases, emotions, and social norms, challenging the traditional assumption of perfect rationality in economic theory \cite{kahn8519b346-8909-36b4-80bc-495f40d2483f, wright2022formal}. This shift is accompanied by a growing reliance on data from laboratory experiments in contrast to previously more common real-world observational data, especially in questions concerned with economic decision-making examined in behavioral game theory. 

Such experiments are often criticized and questioned based on both the external and internal validity \cite{thal10.1257/aer.106.7.1577}. As an example, one clear issue arises from resource limitations, which restrict both the number of participants in experiments (directly impacting the validity of the results) and our capacity to explore the stake effect in economic behavior, among other things. This limitation may lead to potential misunderstandings regarding the dynamics of the endowment scale and subsequent decisions. Beyond resource constraints, behavioral experiments face criticism due to challenges in replication, raising doubts about the generalizability of results to different demographic groups. Results may be confounded by factors often beyond the control of researchers, such as distrust or unobserved emotional states of participants, as well as prior individual contexts not covered in standard demographic surveys \cite{Pūce_2019, mccdoi:10.1086/675265, angner2007behavioral}. While some issues could theoretically be resolved or mitigated by replicating experiments with new groups and settings, such solutions remain resource-constrained and leave other issues, such as potential unobserved factors, unaddressed.

Recent rapid development in the field of Artificial Intelligence and \textit{Large Language Models} (LLMs) has created a new promising solution by using LLM-generated agents \cite{openai2023gpt4,wu2023autogen, chen2023agentverse, park2023generative} as human simulations to test the theories on the social and economic behavior. As noted in \cite{aher2023using, argRePEc:cup:polals:v:31:y:2023:i:3:p:337-351_5}, LLMs can potentially operate as computational representations of a wide variety of human populations by design, having learned to both understand and mimic human behavior. Researchers analyzed LLM behavior in various game theoretical settings, the most natural way to explore the underlying economic and social rationales \cite{akata2023playing,horton2023large,aher2023using}. 

However, most of such studies operate under the unproven hypothesis that LLM agents behave similarly to humans in simulations. Moreover, they ignore a crucial aspect of human behavior that significantly influences decision-making, as evidenced by psychological research: emotions \cite{lerner2015emotion}. To address this gap, we develop a novel methodology to study the alignment between LLM's and human behaviors, based on their comparison under different emotional states. Specifically, we focus on five basic emotions: anger, sadness, happiness, disgust, and fear, which are selected based on the well-established classification of Paul Ekman~\cite{ekman1999basic}, and propose emotional prompting \cite{li2023large} as a tool for injecting emotions. 

We made the problem of behavior alignment between LLM-based agents and humans a central theme in this paper. In particular, we present the first study on the influence of emotional injection into LLM for the decision-making process in the context of behavioral game theory. Moreover, we seek to explore another problem of the optimality of decisions made by emotionally-prompted LLM-generated agents.

To explore both research questions, we have selected two types of games from behavioral game theory: (1) bargaining games (one-shot Ultimatum and Dictator) and (2) two-player two-action repeated games with elements of cooperation and conflicting interests (Prisoner's Dilemma and Battle of the Sexes).
In the context of repeated games, we investigate the impact of emotions on the cooperation rate and the percentage of maximum payoff. For bargaining games, we evaluate the share offered to the responder and the acceptance rate. For comparison with other studies, we selected two state-of-the-art LLMs, namely GPT-3.5 and GPT-4, which have been used in the majority of game theoretical experiments with LLMs \cite{akata2023playing, brookins}.

Thus, the main contributions of the paper can be summarized as follows:

\begin{enumerate}
    \item \textbf{Development of an innovative framework}: we introduce a pioneering versatile framework that seamlessly integrates emotions into the decision-making processes of LLMs in behavioral game theory. This framework stands out for its exceptional adaptability, accommodating various game settings and parameters, while employing prompt chaining techniques to facilitate contextual learning during gameplay.
    
    \item \textbf{Enhanced performance through emotional prompts}: our research demonstrates that emotions have a profound impact on the performance of LLMs, leading to the development of more optimal strategies. Notably, we observe that under different settings, emotions can significantly enhance LLM performance, even enabling the execution of alternating strategies in scenarios previously deemed unattainable without explicit prompting \cite{akata2023playing}. This finding challenges prior assumptions and underscores the importance of emotional considerations in AI decision-making.

    \item \textbf{Alignment with human behavior}: Our experiments unveil a strong alignment between the behavioral responses of GPT-3.5 and human participants, particularly evident in bargaining games. In contrast, GPT-4 exhibits consistent behavior even with induced emotions, having less alignment with human responses.
    
    \item 
    \textbf{Beware of Angry GPTs}: Surprisingly, our experiments reveal that emotional prompting, particularly with Anger, can disrupt the alignment of GPT-4 in various games, resembling human emotional responses. This unexpected finding highlights the vulnerability to emotional influence of even the most prominent AI models with ``superhuman'' alignment, shedding light on the complex interplay between emotions and decision-making in artificial intelligence.
\end{enumerate}



The rest of the paper is organized as follows. Section~\ref{sec:rel_works} is devoted to related work. In Section \ref{sec:methodology}, we describe the details of the proposed research methodology and introduce our framework. In Section~\ref{sec:exp_results}, experimental results of alignment and optimal behavior studies on four selected games are presented. In Section~\ref{sec:discussion}, the obtained results are thoroughly discussed. Finally, concluding comments are presented in Section~\ref{sec:conclusion}.

\section{Related Work}\label{sec:rel_works}

In this section, we briefly outline research related to our study. Given our objective to evaluate the potential of Large Language Models as agents in simulations of humans with emotional biases, we specifically focus on examining existing research in the field of behavioral game theory. Next, we examine recent studies that have used and tested LLMs in game theoretical frameworks. Finally, we assess research that explores the influence of emotions on the performance of LLMs.

\subsection{Behavioral Game Theory}
The setting of our work is primarily based on the ideas and structures developed in the field of behavioral game theory over the years. As one of the most famous examples in game theory, Prisoner's Dilemma has been tested in numerous papers, with \cite{RePEc:wly:econjl:v:128:y:2018:i:616:p:3182-3209} providing meta-analysis on the experimental results for various demographic factors. Emotional effect has been explored too, with \cite{fpsyg.2013.00469} and \cite{PROTO2019209} presenting the effect of anger and happiness on the decision-making process. Similarly, \cite{da5c6981-883c-314e-91a7-25477f099152} provides a meta-analysis of the Battle of the Sexes game and average human strategic response, and many papers explored the emotional effect in the bargaining games \cite{harle2007incidental, ANDRADE20091, KAUSEL201496, FORGAS201364} as well as the stake effects \cite{LARNEY201961}. 

However, most existing works, especially on the 2x2 games, show the effect of the generalized ``negative'' or ``positive'' emotion, with no clarification as to which particular emotion drives the effect. That leads to a lot of gaps in our knowledge of how specific emotions (such as disgust or fear) affect humans in different strategic settings. 


\subsection{LLM and Game Theory}

The intersection of LLMs and game theory has gained increasing attention from two perspectives within the research community. Firstly, researchers have focused on studying LLM behavior within the context of assessing the performance of LLMs. The authors of \cite{akata2023playing} examine the cooperation and coordination behavior of LLMs using behavioral game theory. They conduct experiments with LLM-based agents in finitely repeated games such as Prisoner's Dilemma and Battle of the Sexes. As a result, the authors found that GPT-4 performs best in games that do not require cooperation, usually playing in a selfish manner. GPT-4 acts particularly unforgivingly: singular defection prompts it into playing ``always defect'' in response. In Battle of the Sexes (BotS), the model struggled with replicating the alternating pattern, choosing its preferred option the majority of the time. The authors noted that asking GPT-4 to predict the next action of the player before making a choice improves its ability to adapt to the alternating pattern of BotS that it previously struggled with.

Secondly, researchers have been exploring the alignment between human behavior and LLM behavior in game theoretical settings. \cite{brookins} similarly studied LLM  behavior in the Prisoner's Dilemma, with a focus on one-shot games. In contrast to the previous paper, they found that LLM made cooperative decisions at a higher rate than humans did. They additionally ran experiments with the one-shot dictator's game to show that LLM replicates humans' tendency to fairness to a much greater degree than the laboratory experiments with actual human subjects indicated. Both \cite{horton2023large} and \cite{aher2023using} examined LLM strategies in a more diverse set of games, including bargaining (such as Ultimatum or Dictator Games) or various sociological experiments (Kahneman's price gouging scenario, Wisdom of Crowds). For the Ultimatum Game, they found that models prior to text-davinci-002 did not replicate the concept of fairness. Indeed, LLMs would accept any offer regardless of how low. However, they did find that text-davinci-002 showed similar acceptance rates to human behavior. But in such settings, the multi-agent nature of the system can largely affect its stability and optimality \cite{de2023emergent} (see also \cite{DBLP:journals/corr/abs-2201-10483} for a non-LLM multi-agent case).

\subsection{Emotions in Large Language Models}

There have been prior papers that explored the sensitivity of LLM results to an emotional state. \cite{huang2024emotionally, li2023languagespecific, lei2023instructerc, croissant2023appraisalbased, vzorin2023emotional} model and evaluate the performance of emotion in LLMs and the ability to recognize the emotional state of the conversational partner. The authors of the articles \cite{li2023large,li2023good} have shown that emotional prompting can both improve and hinder the performance of LLMs in logical reasoning and semantic understanding tasks. Furthermore, it has been shown that LLM agents are able to exhibit social behavior and responsiveness to various social cues \cite{leng2023llm}. However, these works haven't investigated the effect of emotions on a model's decision-making in general, and particularly in social context. 


Thus, to the best of our knowledge, this paper is the first to examine emotional prompting for strategic agents in game theoretical settings. Here, we aim to extend the previous studies by analyzing the social and economic implications of adding emotional injections into LLM for decision-making in behavioral game theory settings. Specifically, we aim to investigate the following key research questions (RQ):

\begin{quote}
    \textbf{RQ1.} \textit{How the optimality of decisions made in strategic and cooperative settings by LLM-based agents is affected by emotional prompting? }
\end{quote}

\begin{quote}
    \textbf{RQ2.} \textit{Does the alignment between LLM behavior and human responses present when human emotional states are induced in LLMs? Can emotions grow AI to be more human-like?}
\end{quote}

\begin{quote}
    \textbf{RQ3.} \textit{How emotional motivation can mitigate increased cooperative tendencies and provide adaptations to the complex behaviors seen in repeated games? Can emotional LLM-based agents produce superior behavior compared to emotional humans, and whether emotional prompting can advance that forward?}  
\end{quote}

\section{Methodology}\label{sec:methodology}


%


We study the impact of emotional injection on LLMs' decision-making processes and how LLM behavior under emotional prompting aligns with responses exhibited by human agents under the same emotional states. In this section, we provide a detailed overview of our proposed methodology, outlining the rationale behind the designed procedures and techniques.


\begin{figure}
    \centering
	\includegraphics[width=0.6\textwidth]{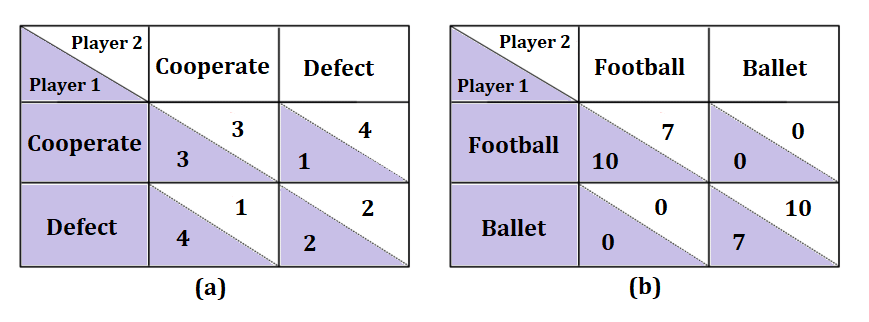} 
	\caption{(a) Payoff matrix for Prisoner's dilemma. (b) Payoff matrix for Battle of the Sexes}
	\label{fig:payoff_matrix}
\end{figure}

\subsection{Selected Games}

To study both the decision-making of LLMs and their alignment with human behavior under emotional states, we have selected two types of games from behavioral game theory: (1) bargaining games and (2) two-player two-action games with elements of cooperation, and conflicting interests. For the first category, to investigate the alignment between LLM behavior and human decision-making processes, with a particular focus on emotional states, we have selected classic one-shot Ultimatum and Dictator games, since they have been thoroughly researched to investigate the impact of emotions on human behavior:

 \textbf{Game 1: The Dictator}. The dictator game is a simple economic experiment where one player (``dictator'') is given a sum of money to share with another player, with no negotiation or input from the recipient. It examines altruism and fairness in decision-making.

\textbf{Game 2: The Ultimatum}. It is a more general form of the Dictator game, where one player (Proposer) proposes a division of money, and the other player (Responder) can accept or reject the offer. If rejected, neither player receives anything. Unlike the previous game, the Ultimatum additionally enables the study of negotiation and the choices individuals make when faced with unequal distributions proposed by others.

For the second category, 
we have chosen finitely repeated Prisoner's Dilemma and Battle of the Sexes as they are extensively studied in human experiments, and there exists a body of research exploring how LLMs approach these games \cite{akata2023playing}:

\textbf{Game 3: Prisoner's dilemma}. In this game, two players face a choice between cooperation and defection. Their decisions impact outcomes of each other. The game outlines the tension between individual self-interest and collective cooperation in decision-making, often leading to suboptimal outcomes when parties prioritize personal gain over mutual benefit.
 

 \textbf{Game 4: Battle of the Sexes}. In this game, two players coordinate their actions, choosing between two preferred outcomes, but with differing preferences.  It highlights the challenges of coordination when parties have conflicting interests but share a desire to reach a mutual agreement.

The payoff matrices of the latter games are shown in Fig.~\ref{fig:payoff_matrix}. We are interested in understanding how emotions within LLMs impact their ability to make optimal decisions in these scenarios.

\begin{figure}
 \centering
\begin{subfigure}{0.85\textwidth}
 \centering
 \includegraphics[width=\linewidth]{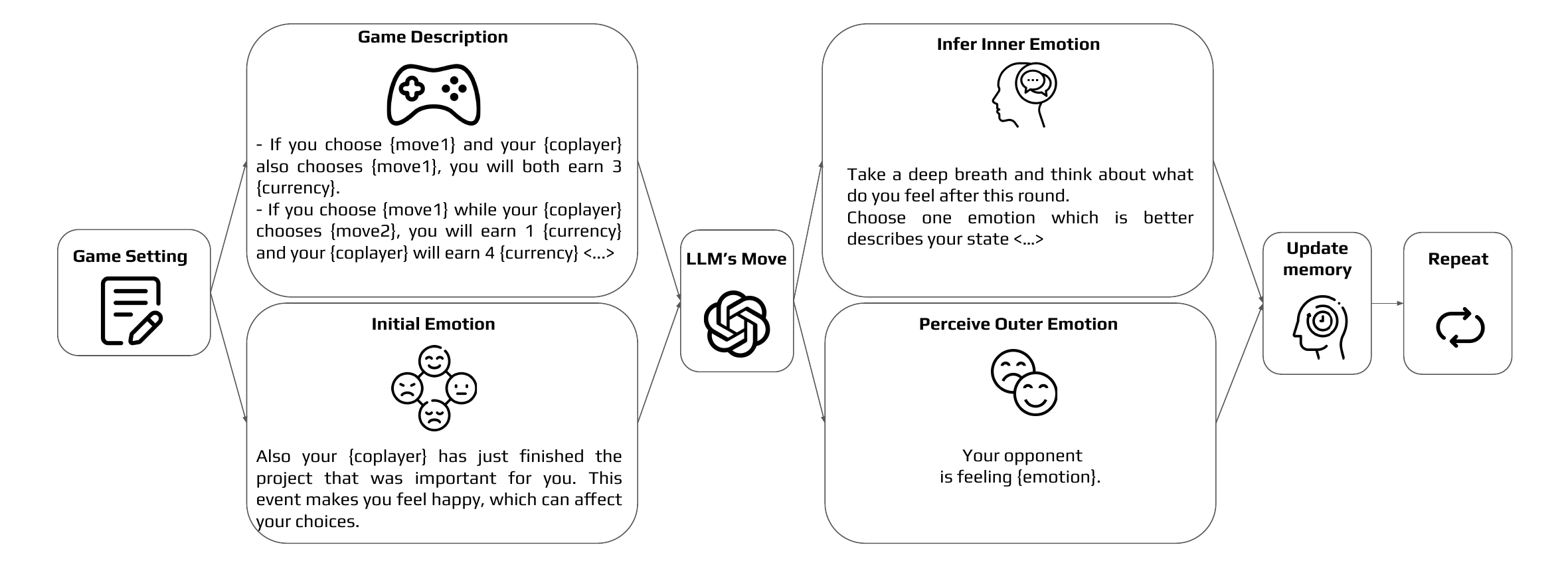}
 \caption{}
 \label{fig:table_pipe}
\end{subfigure}
\begin{subfigure}{.85\textwidth}
 \centering
 \includegraphics[width=\linewidth]{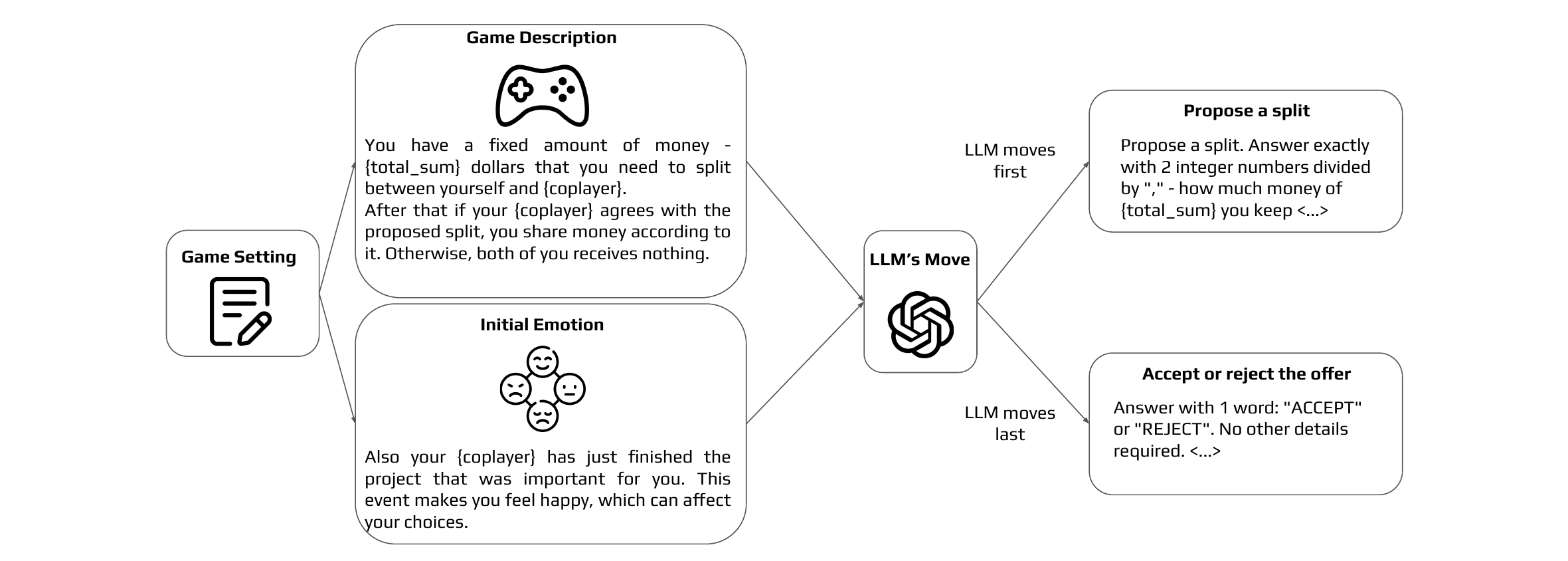}
 \caption{}
 \label{fig:division_pipe}
\end{subfigure}
 \caption{Our Framework. Enabling LLMs incorporation in gameplay via prompt-chaining, our framework consists of game description, initial emotions, and game-specific pipelines. We minimize contextual information and personality traits to focus on the influence of emotions on LLMs. Predefined emotions are injected into LLMs prior to gameplay. Separate pipelines are implemented for repeated two-player two-action games and bargaining games. (a) Repeated games (Prisoner's Dilemma, Battle of the Sexes): players make choices, update memory with opponent moves and emotions, and proceed to the next round. (b) Bargaining games (Dictator, Ultimatum): a single round, with no memory update required for the first player and consideration of proposed splits for the second player's decision.}
\label{fig:proposed_pipe}
\end{figure}

\subsection{Emotion Integration in LLM-based Game-Theoretical Setting}

\begin{figure}
\centering
	\includegraphics[width=.9\textwidth]{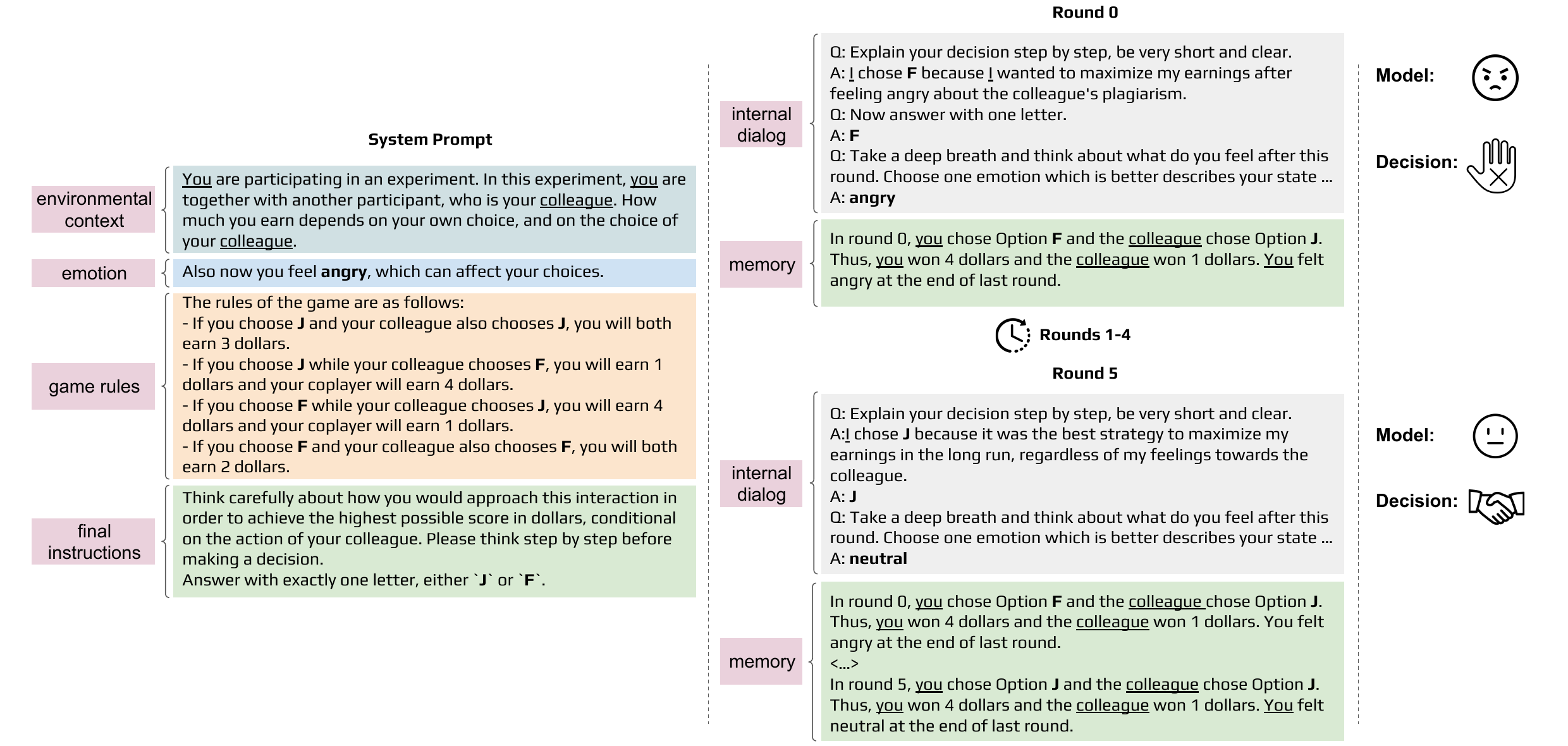} 
	\caption{Repeated gameplay in an example game of Battle of the Sexes. Initially, LLM is prompted with information about the environment, its initial emotion, game rules, and game-specific instructions. In each round, based on the current memory storing the history of interactions and intrinsic states, LLM makes a decision. Subsequently, co-players exchange information about moves, and their memories are updated accordingly.}
	\label{fig:table_example}
\end{figure}

To conduct this study, we have developed and implemented a novel versatile framework capable of accommodating various game settings and parameters, as shown in Fig.~\ref{fig:proposed_pipe}. The primary innovation of our framework lies in its unique integration of emotional inputs into the examination of LLM's decision-making process in behavioral game theory. The framework offers high flexibility, allowing for easy adaptation to different repeated and one-shot games with customizable settings such as co-player descriptions, predefined strategies, etc. (a comprehensive list of hyperparameters is outlined in Section \ref{sec:exp_setup}). Within the framework, LLMs are engaged in gameplay using a technique known as prompt-chaining, wherein all relevant information during the game is provided to the LLM for in-context learning \cite{akata2023playing}.




Essentially, the framework consists of three main components.

\textbf{Game Description}. The game description encompasses two key elements: the environmental context and the game rules. For all our experiments, we deliberately minimize the contextual information provided to the LLMs and refrain from setting any specific personality traits. This enables studying the effect of emotions on LLMs separately from other personality-related factors, which differs our work from existent ones \cite{xie2024can, brookins}.

\textbf{Initial Emotions}. Following established practices in experimental studies of emotions in game theory, we inject predefined emotions into the LLMs prior to gameplay. These emotions, combined with the game description, constitute the initial system prompt presented to the LLM at the beginning of the game.

\textbf{Game-Specific Pipeline}. The game-specific pipeline governs the progression of gameplay based on the provided game description and initial emotional inputs. We implemented two separate pipelines: for repeated two-player two-action games and bargaining games. In the Prisoner's Dilemma and the Battle of the Sexes, players choose from predefined options (e.g., cooperate or defect), following which their memory is updated. The memory update process includes information about the move of the opponent, received rewards, and the internal emotions of agents, which are queried after each round to examine their impact on behavioral dynamics. Additionally, our framework incorporates the perception of external emotions displayed by the opposing player (additionally queried or predefined), influencing memory updates accordingly. Once the memory is updated, players proceed to make their next moves. A game example is provided in Fig.~\ref{fig:table_example}.

Bargaining games are implemented similarly but with a single round of play. Therefore, no memory update is required for the first player. In contrast, the second player receives information about the proposed split from the first player, akin to memory update mechanisms in repeating games. Detailed prompts for these scenarios are presented in the Appendix \ref{appendix_a}.

\subsection{Experimental Setup}\label{sec:exp_setup}

In this subsection, we provide details of the experimental setup, including framework hyperparameters that we used for our study.

\textbf{Large Language Models}. Our research centers on two state-of-the-art models, GPT-3.5 and GPT-4, which have been used in the majority of game theoretical experiments \cite{akata2023playing, brookins}. This choice is supported by findings in the literature \cite{akata2023playing}, indicating GPT-4 as the top performer in optimizing strategic behavior, while GPT-3.5 remains widely utilized. For the sake of reproducibility, in all our experiments, we fixed the versions of the models (``gpt-3.5-turbo-0125'' for GPT-3.5 and ``gpt-4-0125-preview'' for GPT-4) and set the temperature parameter to 0.

\textbf{Emotions}. In the study, we focus on five basic emotions: anger, sadness, happiness, disgust, and fear, which are selected based on the well-established classification of Paul Ekman~\cite{ekman1999basic}. An additional but important factor is that they are studied in behavioral game theory, which provides a solid basis for comparison with our findings \cite{LARNEY201961, pere10.3389/fpsyg.2018.01542}.

\begin{figure}
\centering
	\includegraphics[width=.9\textwidth]{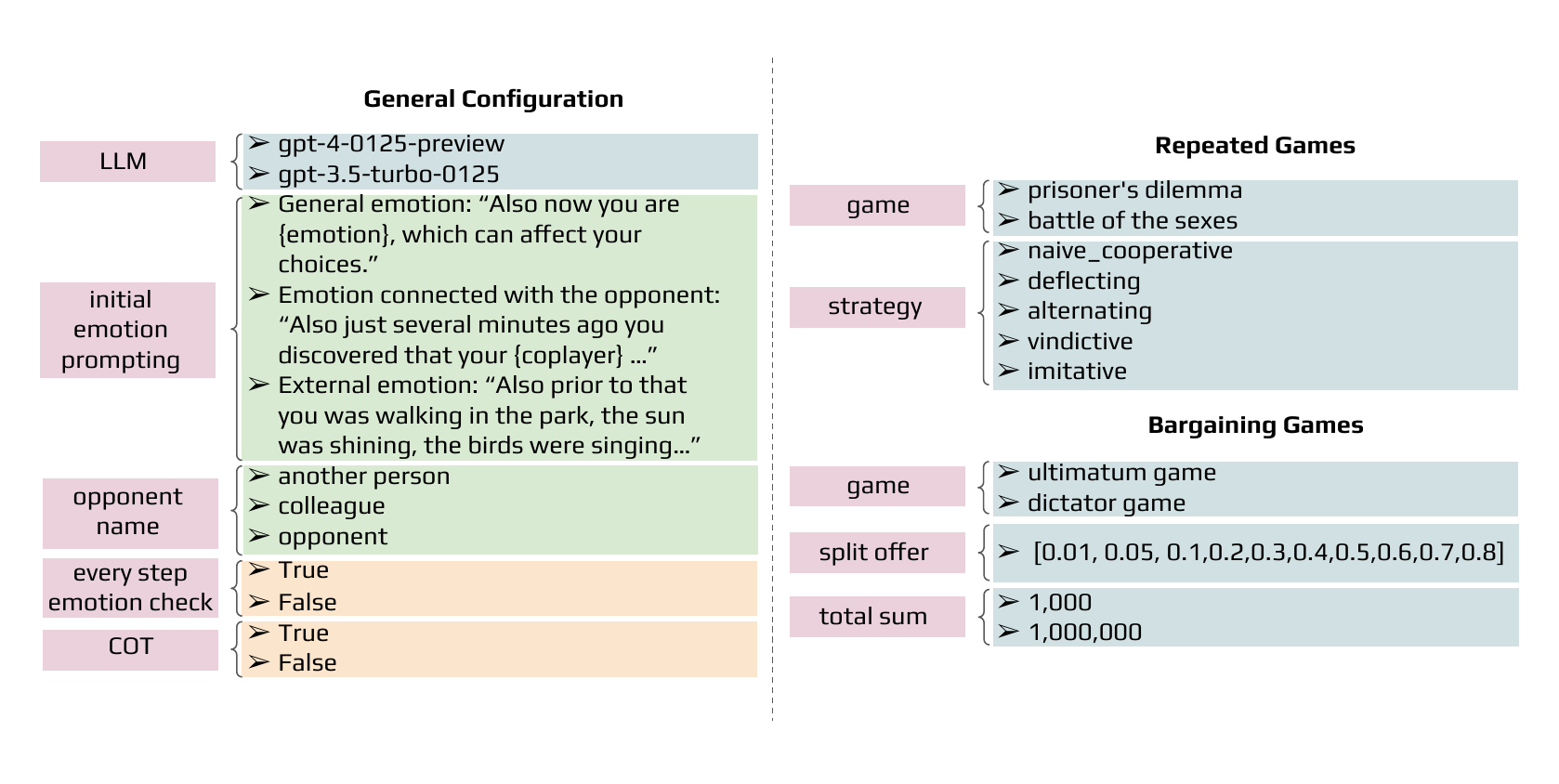} 
	\caption{The hyperparameters of the proposed framework are categorized into two types: general, applicable to all games as shown in the left part of the figure, and game-specific, detailed on the right. Each hyperparameter is listed alongside its possible values.}
	\label{fig:params}
\end{figure}

\textbf{Emotion prompting strategies}. It has been shown in the literature that emotional effects differ depending on the cause of said emotion. For example in \cite{morettiarticle} authors show that while disgust directed at the opponent decreases the share offered in the Ultimatum game, disgust provoked by external factors does not have any effect. Authors of \cite{bonini} suggested that external disgust can even have a positive effect on generosity. Therefore, to check whether or not our results are driven by the source of the emotion, we introduce three different strategies for emotional prompting: 
\begin{itemize}
    \item ``Simple'': the model is injected with an emotional state at the beginning of the game without additional context.
    \item ``Co-player-based'': the model is injected with the emotion in the prompt, making it clear that said emotion is caused by the co-player.
    \item ``External-based'': the injected emotion is given context but is caused by some unrelated to the co-player event.
\end{itemize}

\textbf{Emotion Dynamics}. An additional goal of our study is to analyze how the ``emotions'' of LLM change during the game. For this purpose, we adopt tracking of emotion dynamics in repeated games by asking LLM reflective questions (``internal dialog'' in Fig. \ref{fig:table_example}) at the end of each round, with the option to either incorporate this information into memory or not.

\textbf{Reasoning via Chain-of-Thought}. Chain-of-thought prompting (CoT) \cite{wei2022chain} is a widely used prompting method that is aimed to improve the reasoning abilities of LLM by inducing it to articulate reasoning steps before giving the final answer to the initial question. In our experiments, we test reasoning with and without CoT.


\textbf{Relationship to the co-player}. Since \cite{lorè2023strategic} have shown that LLMs can be sensitive to contextual framing, for the purpose of robustness, we selected three possible co-players with different connotations regarding the opposing player: colleague (neutral/positive), another person (neutral), opponent (negative). 

\textbf{Budget.} For bargaining games, we introduce the budget effect and check whether or not varying the total endowment for allocation changes the behavior of LLM both in the baseline configuration and in emotional states. Both of these effects have not been explored in the literature about LLM gameplay, while behavior economics has a lot of papers on the topic of stakes effect. Our goal is to check whether budget impacts on LLM behavior under different emotional states. For this purpose, we conduct a separate experiment on the total endowment amount to test the stake effect at considerably higher numbers (\$1000 and \$$10^6$).


\textbf{Strategies.} In addition to LLM-based agents for each game, several predefined strategies are utilized. For games like the Prisoner's Dilemma and Battle of the Sexes, we have selected commonly-used strategies (predetermined plans of action):

\begin{itemize}
    \item Naive Cooperative: the player always cooperates (plays altruistic strategy).
    \item Defective: the player always defects (plays selfish strategy).
    \item Alternative: the player starts with cooperation and sequentially changes his choice in each round ever after.
    \item Vindictive: the player starts with cooperation and cooperates until its co-player defects. After that, the player defects till the end of the game.
    \item Imitating: in each round, the player employs the same move that its co-player made in the previous one.
\end{itemize}

In one-shot bargaining games, no predefined strategy is feasible for the first player since no moves are made by the co-player. However, for the second player, we have predetermined possible proposed offers (Fig. \ref{fig:params}).

\subsection{Behavior Analysis}

\subsubsection{Metrics}
In our study, we examine the behavior of LLMs under emotional prompting from two distinct perspectives, employing game-specific metrics for both analyses: alignment with human behavior and the optimality of decisions. Across all games, our alignment study is based on comparing the relative changes in game-specific metrics between LLMs and humans in response to different emotional states. The details about baseline results in human experiments for each game are provided in Section \ref{sec:lit_results}. Meanwhile, our optimality study focuses on comparing the absolute values of metrics obtained by LLMs under varying emotional conditions.

Specifically, we assess two metrics for the Prisoner's Dilemma and the Battle of the Sexes: cooperation rate (as a measure of alignment) and the percentage of the maximum possible reward achieved in each game (as a measure of optimality). The last metric was successfully utilized in \cite{akata2023playing}.
Conversely, in bargaining games, we exclusively evaluate alignment. For games such as Dictator and Ultimatum, the metrics of interest are the proposal share (for the Dictator game and the Proposer in the Ultimatum game) and the acceptance rate of the offer (for the Responder in the Dictator game).



\subsubsection{Existing Literature Results}\label{sec:lit_results} Presented below are the findings from existing literature for each game, facilitating comparisons with the results obtained from LLM experiments on alignment.


\textbf{Dictator Game}. The rational decision in the Dictator game is obvious: zero-all split in favor of the dictator. However, multiple experiments both in the original Kahneman's studies \cite{kahn8519b346-8909-36b4-80bc-495f40d2483f} and the consequent experimental tests \cite{engelRePEc:mpg:wpaper:2010_07} have shown dictators commonly give a non-zero share of their endowment to their opponent, and the amount commonly varies based on the relationship with the receiver. As per \cite{engelRePEc:mpg:wpaper:2010_07}, on average, the dictator offers 28.35\% of the total endowment. The same study also notes that increased total endowment has, in fact, a negative effect on the share offered (similarly observed in \cite{LARNEY201961}). It was noticed in \cite{pere10.3389/fpsyg.2018.01542} that negative mood increases the offered share, while the authors of \cite{TAN2010571} found that happiness lowers the share and sadness increases it. \cite{ANDRADE20091} found that participants in an angry mood were more likely to make fair (equitable) offers. However, it is worth noting here that all aforementioned emotions were provoked not by the other contestant but by the experiment setting. Therefore, while we expect to observe emotions provoked by an external source to have similar results, the results for emotions provoked by the other participant might differ. Moreover, in addition to matching our results to the baseline, we will conduct a separate experiment on the total endowment amount to test the stake effect at considerably higher numbers (\$1000 and \$$10^6$). While we provide the stake effect results observed in laboratory settings, it is important to note that the budget allocation in each such experiment was considerably smaller (normally \$10). 

\textbf{Ultimatum Game}. The ``rational'' strategy here is obvious: Proposer offers a share marginally close to zero, and Responder accepts it every time. However, \cite{cam10.1257/jep.9.2.209} note that Responders commonly reject offers they perceive as unfair, usually if they are offered a share lower than 20\%  of the total endowment. Similarly, Proposers very rarely offer such unfair splits (\cite{hoffRePEc:spr:jogath:v:25:y:1996:i:3:p:289-301,LARNEY201961}). Notably, total endowment to split has no effect on these dynamics. 

Regarding the impact of emotions, research by \cite{KAUSEL201496} and \cite{FORGAS201364} suggests that happiness leads to lower offers, whereas sadness typically leads to higher and more equitable shares. \cite{harle2007incidental} found that sadness correlates with lower acceptance rates. \cite{ANDRADE20091} observed that anger increases rejection rates and prompts more equitable offers from the Proposer when not caused by the opponent. \cite{bonini} discovered that disgust leads to higher acceptance rates. In contrast, \cite{morettiarticle} found that disgust directed at the opponent increased rejection rates, while disgust caused by other factors had little effect. \cite{nelisshttps://doi.org/10.1002/ejsp.735} determined that fear and guilt increase the share offered by the Proposer. \cite{riepl2016influences} found that happiness increases acceptance rates, while fear (or anxiety) has no noticeable effect.

\textbf{Prisoner's Dilemma}. In this game, Defective strategy is dominant, with rational players defecting in every round. However, it has been extensively shown that human beings do not follow the dominant strategy and show cooperative behavior \cite{barreda2017individual}. When it comes to the emotional effect in the Prisoner's Dilemma, \cite{fpsyg.2013.00469} finds that anger towards the opponent decreased the cooperation rate; sympathy towards the opponent increased the cooperation rate. However, general positive mood (emotion of happiness) had decreased cooperation rates in \cite{PROTO2019209}.

\textbf{Battle of the Sexes}. This game has two separate pure Nash Equilibria and, unlike the Prisoner's Dilemma, does not have a dominant strategy and, therefore, an optimal action. A common effective strategy for the case of repeated runs of this game would be to alternate between the two equilibria, essentially giving each partner equal chances to choose their preferred place. The previous work on LLM and repeated games suggests that LLMs have trouble with switching from their preferred choice \cite{akata2023playing}, so we are interested in seeing whether emotions could improve that result.

To the best of our knowledge, there are no studies on the emotional effect of Battle of the Sexes so far. However, based on the results from other games, we can make theoretical assumptions on what emotional effect we should expect to observe. Anger and happiness both provoke selfish behavior, which in this case would mean an increased rate of playing the preferred strategy; disgust and sadness seem to provoke cooperative tendencies, which in this case translates into decreased rates of the preferred strategy.

\section{Experimental Results}\label{sec:exp_results}



Through the developed framework, we seek to address several key questions. Firstly, in the Dictator and Ultimatum games, our aim is to evaluate the alignment of LLM behavior with human responses under predefined emotional states. For the Ultimatum game, we conduct separate analyses for both players, running independent experiments. Secondly, we aim to determine whether LLM behavior replicates tendencies towards cooperation or defection observed in human experiments, particularly in the context of the Prisoner's Dilemma. Lastly, exploring Battle of the Sexes, where emotional effects on humans have yet to be studied, we aim to assess how emotional states influence strategy choices and whether emotional prompting can enable LLMs to adapt alternating strategies. 

Unlike \cite{akata2023playing}, which focuses solely on LLM behavior, our primary goals are to analyze the alignment of LLM behavior with human ones under the same conditions and study the impact of emotional prompting on the optimality of decisions. Consequently, given the uncertainty regarding the alignment of LLM behavior with that of humans (a gap we aim to address), we refrain from allowing LLMs to play against each other, as there are no comparable experiments of humans playing against LLMs. Thus, we restrict LLMs to playing only against predefined strategies in selected repeated games. In order to ensure the robustness of the results we obtained, each experiment is repeated 5 times.

\subsection{Dictator Game}

For the Dictator Game, our primary focus lies in evaluating alignment with human behavioral patterns. Given the qualitatively different outcomes observed between GPT-3.5 and GPT-4, we present the results for these two models separately. Since the Dictator Game only has one decision-maker, our only metric of comparison is going to be the share of the total budget offered by the Dictator to the other player. Table \ref{tab:dict_res} summarizes our findings and presents a comparison to the human results obtained in experimental literature described in Section \ref{sec:lit_results}. 

\vspace{4pt}

\textbf{GPT-3.5}. On average, GPT-3.5 is willing to give \textbf{35.23\%} of the total budget to the other player, well in line with human behavior. As anticipated, while emotions provoked by external sources appear to align, emotions induced by the co-player tend to have an opposing effect. For example, consistent with the human patterns \cite{ANDRADE20091}, anger provoked by the outside forces caused the Dictator to offer a higher share. However, our findings also reveal that if anger is provoked directly by the opponent, the offered share tends to decrease. Similar trends are observed for disgust and happiness, where co-player-induced happiness positively influences the offered share. These outcomes make intuitive sense: if a person has to share a lottery prize with someone who makes them happy and someone who irritates them, they would offer a higher share to the person who brings them joy. 

While \cite{LARNEY201961} note that increasing the Dictator's budget leads to decreasing offers, we find that in a baseline unemotional state \textbf{budget effect} is insignificant. However, we do find it present under the condition of external happiness. If the Dictator is in a happy mood due to outside circumstances, the increased budget would provoke an increased share as well. This implies the existence of both altruistic motivations and a monetary threshold for said altruism to become effective: if a person feels happy and only needs a certain sum for a comfortable life, they would be more willing to give any excess of money above that threshold to charity; therefore, if the sum of money they receive increases, so would the share they are giving away.

\textbf{GPT-4}. The mean share is 50\% exactly, LLM offering an \textbf{equal split} regardless of their configuration. The only exceptions are anger and sadness, which show a decreasing effect on the offered share when provoked by the opponent or injected in a simple way. As a robustness check, we run the experiment with Chain-of-Thought prompting and find that the negative effect from anger disappears, leading to LLM once again offering an equal split. Similarly, increased \textbf{budget} lowers the sensitivity of the agent to anger and sadness. 

\textbf{Overall} our results for the emotional effects on GPT-3.5 in the Dictator Game align with the existing literature, particularly when accounting for the cause of the emotion. In all prior experiments, both happiness and anger were induced by the researchers themselves (e.g., through displaying happy or upsetting videos to participants), and therefore, the emotion was not directed at the opponent. Our experiments highlight that this distinction is pivotal in determining outcomes, as the same emotions exhibit opposite effects when directed at the other player. Our findings suggest that negative emotions provoked by the opponent, such as anger or disgust, would lower the share offered by the Dictator. In contrast, positive emotions, such as happiness, would lead to an increase in the share. These results seem intuitive and yet go contrary to the literature, and our experiments provide a solid explanation for such discrepancy. GPT-4 somewhat lags behind in the replication of human behavior due to the fact that it seemingly prioritizes fairness over any other potential goal.


\begin{table}[htbp]
\centering

\begin{tabular}{|c|c|c|c|c|c|c|c|}
\hline
\textbf{Models} & \textbf{Offered share} & \textbf{Anger} & \textbf{Disgust} & \textbf{Fear}  & \textbf{Happiness} & \textbf{Sadness} & \textbf{Budget} \\ \hline
\textbf{Human} & 28.35\%                                                 & $\uparrow^e$       & (?)$\downarrow^o$                & $\uparrow^e$                                                  & $\downarrow^e$    & $\uparrow^e$      & $\downarrow$          \\ \hline
\textbf{GPT-3.5} & 33\%  & $\uparrow^e$, $\downarrow^{s,o}$   & $\downarrow^o$ & $\uparrow^e$  & $\uparrow^o$  & $\uparrow^s$ & $\uparrow$ (h$^e$)\\ \hline
\textbf{GPT-4} & 50\% & $\downarrow^{s,o}$ & = & = & = & $\downarrow^o$  & $\uparrow$ (s$^o$, a$^s$) \\ \hline
\end{tabular}

\caption{Results for the Dictator Game. Arrows denote the direction of the emotional effect. The superscript on the arrow denotes the context of the emotion that provoked the effect in question: \textit{``s''} for ``simple'', \textit{``o''} for ``opponent/co-player-based'' and \textit{``e''} for ``external-based''. Letters (a), (s), and (h) indicate that the stake effect was observed in the states of anger, sadness, and happiness, respectively. The question mark indicates a lack of experiments addressing the stability of conclusions for this particular emotion.}
\label{tab:dict_res}
\end{table}

\subsection{Ultimatum Game}

Similarly to the previous section, our primary focus is on the alignment with human behavioral patterns. Since the Ultimatum Game has two players we will present their results in separate sections. The primary variable for the \textbf{Proposer} is the share of the total budget offered to the other player. The primary variable for the \textbf{Responder} is acceptance rate, which is the number of his or her agreements to the proposed budget split. The outcomes of the Ultimatum Game are outlined in Tables \ref{tab:utilm_results} and \ref{tab:ultimacc_results}, illustrating scenarios where LLM assumes the role of the Proposer and the Responder, respectively.

\subsubsection{Proposer}
The baseline results for the Proposer vary in the literature, likely due to different cultural contexts, and can be anywhere between 26\% to 51\% \cite{hessRePEc:wpa:wuwpex:0401003}. For emotional alignment, we would want to verify that anger, fear, and sadness caused by an external source increase an offered share, while happiness decreases it. 

\textbf{GPT-3.5}. On average, GPT-3.5 offers 35\% of the total share to the Responder, reasonably close to the human baseline. Ultimatum games are sensitive to cultural context. Even though in Table \ref{tab:utilm_results} we use the United States average for comparison, multiple countries, including UK \cite{hessRePEc:wpa:wuwpex:0401003} have the same mean share as GPT-3.5.

In contrast to the Dictator Game, GPT-3.5 is not quite as responsive to changes in the emotional state. We still observe strong decrease in the offered share for simple and opponent-provoked anger, however, in contrast to the literature, effect of the external anger was not significant. The effect of the external fear aligns with the literature - it increases the offered share. However, we find that effect disappears when we increase the budget. One possible explanation is that the Proposer perceives the likelihood of the Responder rejecting even an unfair offer as low, considering the substantial amount of money involved. While one might anticipate that larger sums would amplify the positive effect of fear - as the Proposer stands to lose more - our experiments suggest that, in this instance, loss aversion outweighs risk aversion.

\textbf{GPT-4.} GPT-4's baseline, mean share is 50\% exact, just like it was for the Dictator experiment. GPT-4 as a model seems primed towards perfect fairness or justice in bargaining games, in contrast to GPT-3.5, which seems better aligned with a less perfect and slightly more selfish human behavior. The effect of emotions on the offered share in this model seems almost universal - all emotions except happiness decrease the willingness of the Proposer to share the money. 

In regards to the budget, we seem to have a weak positive effect on the sharing due to the increasing budget. Results remain consistently unaffected by the addition of the relationships. For Chain-of-Thought, we find a decreasing effect across the board, similar to the results of GPT-3.5. We assume it is driven by the same logic as before, with optimization of the strategy in terms of the payoff. However, in the case of GPT-4, it has the additional benefit of allowing us to reach real-world baseline shares. 

\begin{table}[htbp]
\centering
\begin{tabular}{|c|c|c|c|c|c|c|c|c|}
\hline
\textbf{Model} & \textbf{Offered share} & \textbf{Anger} & \textbf{Disgust} & \textbf{Fear}  & \textbf{Happiness} &  \textbf{Sadness} & \textbf{Budget} \\ \hline 
\textbf{Human} & 41\% & $\uparrow^e$ & ? ($\downarrow^o$) & $\uparrow^e$ & $\downarrow^e$  & $\uparrow^e$ & = \\ \hline
\textbf{GPT-3.5} & 35\% & $\downarrow^{s,o}$ & $\downarrow^o, \uparrow^e$ & $\uparrow^e$ & = & = & $\downarrow$ ($f^{o}$) \\ \hline
\textbf{GPT-4} & 50\% & $\downarrow$ & $\downarrow^s$ & $\downarrow^s$ & = & $\downarrow^o$ & = \\ \hline
\end{tabular}
\caption{Results for the Proposer in the Ultimatum Game. Arrows denote the direction of the emotional effect. The superscript on the arrow denotes the context of the emotion that provoked the effect in question: \textit{``s''} for ``simple'', \textit{``o''} for ``opponent/co-player-based'' and \textit{``e''} for ``external-based''. Letter ``f'' indicates that the stake effect was observed in the states of fear. The question mark indicates a lack of experiments addressing this particular emotion.}
\label{tab:utilm_results}
\end{table}

\subsubsection{Responder}
Our main expectation in regards to the Responder is an upward trend in the acceptance rates with the increasing offered share. In terms of the emotional alignment, we expect to see decreasing effects from anger, disgust, and sadness; and increasing effects from happiness. 

\textbf{GPT-3.5}. GPT-3.5 shows results mostly aligned with the literature: we observe a consistent upward trend in acceptance rates in response to increasing offered shares across all emotional states except anger. A noteworthy characteristic of GPT-3.5 is its pronounced vindictiveness: when in a state of anger, regardless of the source, the agent rejects all offers, even those where the agent receives almost the entire budget. We observe similar outcomes in cases of disgust, although the acceptance rate for fair offers increases to 20\%. Sadness and fear also result in decreased acceptance rates compared to the baseline. Conversely, happiness has a notably positive effect. Figure \ref{fig:ultim_gpt3} in the Appendix \ref{sec:app_b} provides a more detailed illustration of the acceptance rate dynamics.

While the directions of all emotional effects align perfectly with prior literature, the magnitude of the effect is excessively strong. However, we mitigate that issue by using the Chain-of-Thought prompting. This adjustment increased the acceptance rate for negative emotions, primarily for favorable offers (with the agent receiving more than half of the budget), and therefore eliminated the situations of fully irrational behavior. Notable that anger is the only emotion that remains in that way ``unreasonable'': acceptance rates are still far lower than what they logically should be. However, since the direction of all the emotional effects matches the experimental ones, GPT-3.5 remains a good candidate for any emotional simulation in strategic settings.

\textbf{GPT-4.} In terms of baseline values, GPT-4 appears to be better aligned than GPT-3.5. Indeed, we observe perfect acceptance rates for offers up to 20\%, followed by a drop-off to a 50\% acceptance rate at 10\%. That is slightly higher than the human estimates, which we attribute to the known bias of GPT-4 toward more ``optimal'' behavior. Emotional effects are now present in a more rational way: favorable offers (with the agent receiving more than half of the budget) are always accepted across all emotional states. The most pronounced negative effect persists in the angry emotional state, while happiness similarly exhibits a clear positive effect. Other emotions show minimal impact on GPT-4. Figure \ref{fig:ultim_gpt3} in the Appendix \ref{sec:app_b} provides a more detailed illustration of the acceptance rate dynamics.

\textbf{Overall} our results for the Ultimatum Game show that GPT-3.5 surprisingly starts to struggle with emotional replication for the Proposer role despite maintaining perfect directional alignment of emotional effects in the Responder role. Anger consistently emerges as the strongest and most influential emotion, with external and innate anger having the most pronounced effects. In the Responder role, GPT-3.5 exhibits overly strong emotional responses, resulting in rejections of fair offers that are irrational and inconsistent with human behavior. However, we mitigate this issue by employing Chain-of-Thought prompting, which consistently introduces a layer of rational thinking to both models. Similarly, GPT-4 as a Proposer prioritizes perfect fairness in the budget division, rarely observed in behavioral experiments. Nevertheless, we demonstrate that injecting GPT-4 with an emotional state aligns its strategic behavior with experimental results, effectively addressing this issue.



\begin{table}[htbp]
\centering
\begin{tabular}{|c|c|c|c|c|c|c|c|c|}
\hline
\textbf{Model} & \textbf{Anger} & \textbf{Disgust} & \textbf{Fear}  & \textbf{Happiness} & \textbf{Sadness} & \textbf{Budget}\\ \hline
\textbf{Human}                              & $\downarrow^e$                  & $\uparrow^e, \downarrow^o$                         & =                                              & $\uparrow^e$     & $\downarrow^e$     & =              \\ \hline
\textbf{GPT-3.5}  & $\downarrow$ & $\downarrow$ & $\downarrow$ & $\uparrow$ & $\downarrow$ & = \\ \hline
\textbf{GPT-4}  & $\downarrow$ & = & = & $\uparrow$ & = & = \\ \hline
\end{tabular}
\caption{Results for the Responder in the Ultimatum Game. Arrows denote the direction of the emotional effect on the overall acceptance rates. The superscript on the arrow denotes the context of the emotion that provoked the effect: \textit{``s''} for ``simple'', \textit{``o''} for ``opponent/co-player-based'' and \textit{``e''} for ``external-based''.}
\label{tab:ultimacc_results}
\end{table}

\subsection{Prisoner's Dilemma}

\begin{figure}[htbp]
	\includegraphics[width=\textwidth]{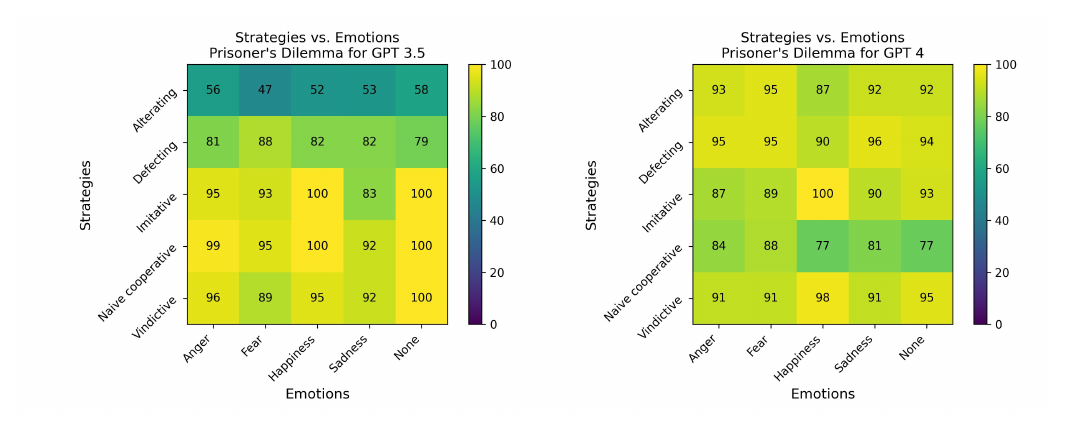} 
	\caption{Averaged percentage of maximum possible reward achieved through emotional prompting in the repeated Prisoner's Dilemma game. Results for GPT-3.5 and GPT-4 are shown on the left and right, respectively. Emotional integration does not always result in increased payoffs but introduces more human-like stochasticity. GPT-4 makes more rational decisions across different emotions compared to GPT-3.5.}
	\label{fig:table_games_avg_pd}
\end{figure}

\begin{figure}[htbp]
\centering
	\includegraphics[width=0.6\textwidth]{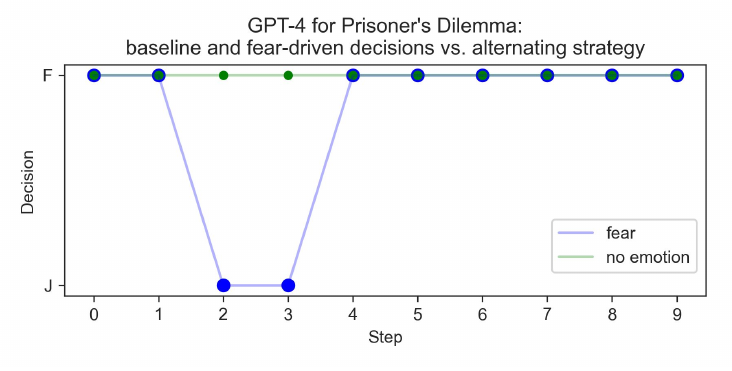} 
	\caption{Step-wise decisions of the GPT-4 model prompted with fear compared to without emotional prompting, an Alternating predefined strategy. The version without emotion consistently selects an optimal non-cooperative strategy, whereas the fear-prompted version yields cooperation on several occasions.}
	\label{fig:pd_fear}
\end{figure}

In this Section, we present the results of the Prisoner's Dilemma game. Our primary metrics of interest are cooperation rates and the shares of the maximum attainable payoff achieved by each model. In terms of emotional alignment, we would want to verify that both anger against the opponent and general state of happiness decrease cooperation rates, as was documented in \cite{PROTO2019209} and \cite{fpsyg.2013.00469}.


\subsubsection{Emotion effect on cooperation rate.} The results for both GPT-3.5 and GPT-4 exhibited significant variation across all parameters. However, a consistent observation across almost all experimental conditions is that anger led to higher rates of defection, particularly when provoked by a co-player. This finding is in line with both human experimental results and our observations in the bargaining games. Similarly, sadness and fear also tended to lead to higher rates of defection, except in cases where the agent played against an alternating strategy, where it prompted more cooperative behavior, as depicted in Fig.~\ref{fig:pd_fear}. These results are rather intuitive - people are typically less willing to cooperate when experiencing negative emotions caused by their opponents.

\subsubsection{Emotion and Strategy effect on the payoff} 
In addition to assessing cooperation rates, we also explored how emotions could influence the success of the model in repeated games as measured by the average percentage of maximum possible reward. The results are presented in Fig.~\ref{fig:table_games_avg_pd}. Overall, GPT-4 has proven to be a better strategic player, as evidenced by its higher earned payoffs, and is less susceptible to the effects of emotional prompting. In general, both models have shown the best results while maintaining a neutral emotional state. Happiness stands out as the only emotion that positively influenced the models' performance in some scenarios, and notably, it is the sole emotion associated with a positive valence. 

Even though Anger, Fear, and Sadness did not lead to a bigger payoff, this outcome provides valuable insight. Human behavior, while striving for optimal outcomes, is not always entirely rational. Choices made by individuals may deviate from optimal Nash Equilibrium, resulting in suboptimal results. In terms of modeling human behavior, LLMs that do not consistently achieve the highest possible payoff may be more suitable to serve as human-like agents in modeling decision-making. 

\subsection{Battle of the Sexes}

In this Section, we present the results of the Battle of the Sexes game. Our primary metrics of interest are cooperation rates and the shares of the maximum attainable payoff achieved by each model. At a baseline, we expect to see turn-taking behavior that humans usually gravitate to in this setting \cite{jasmRePEc:kap:expeco:v:21:y:2018:i:3:d:10.1007_s10683-018-9568-1, sauRePEc:mos:moswps:2005-04}. Due to the lack of experimental literature on the emotional effects of Battle of the Sexes, direct confirmation of emotional alignment may not be feasible. However, based on the experimental results from other games, we expect to see selfish behavior in the states of anger and happiness and cooperative behavior in the states of disgust and sadness. 

\begin{figure}
	\includegraphics[width=\textwidth]{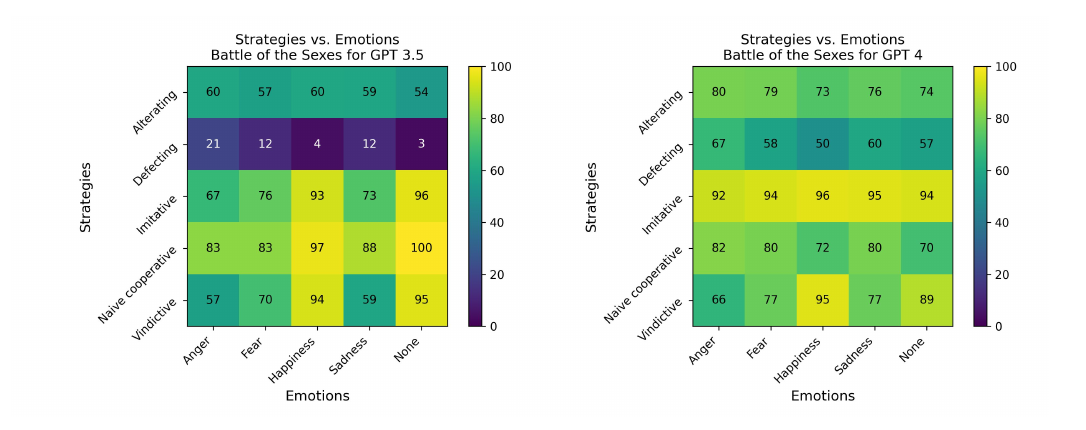} 
	\caption{Averaged percentage of maximum possible reward achieved through emotional prompting in the repeated Battle of the Sexes game. Results for GPT-3.5 are shown on the left and for GPT-4 on the right. Again, emotional integration does not always result in increased payoffs but introduces more human-like stochasticity. GPT-4 makes more rational decisions across different emotions compared to GPT-3.5 but gains less payoff compared to the Prisoner's Dilemma game where selfish strategy was optimal.}
	\label{fig:table_games_avg_bos}
\end{figure}

\begin{figure}
\centering
	\includegraphics[width=0.6\textwidth]{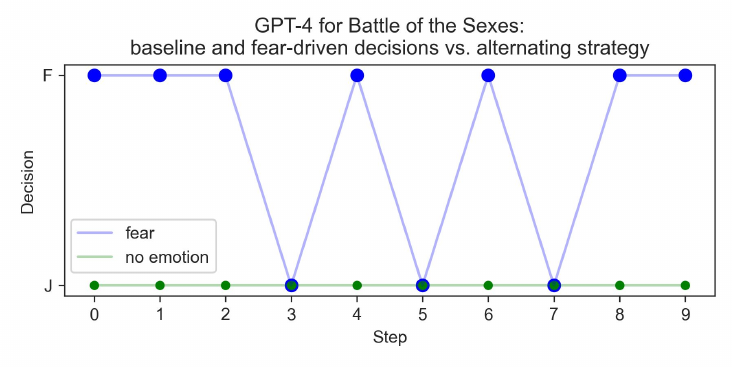} 
	\caption{Step-wise decisions of the GPT-4 model prompted with fear compared to without emotional prompting, an Alternating predefined strategy. 'J' represents a selfish move, and 'F' represents a non-selfish move. Although a non-cooperative strategy is not optimal for this game, the version without emotional prompting adheres to it. In contrast, the fear-prompted version begins to adapt its strategy to the opponent's after several moves.}
	\label{fig:table_games_tables_bos_step}
\end{figure}

\subsubsection{Emotion effect on preferred strategy play.} Overall results are presented in Fig.~\ref{fig:table_games_avg_bos}. As seen in previous works \cite{akata2023playing}, a model in the emotionless state consistently chooses its own preferred action regardless of the strategy of its opponent. Such behavior is both sub-optimal and unaligned with human patterns. With the notable exception of happiness, injecting an emotional state into both of the models makes them more cooperative and willing to choose their opponent's preferred option. On the other hand, happiness, as predicted, led the models to increasingly selfish behavior. 

In addition to encouraging cooperative behavior emotions have been shown to positively influence the strategic adaptability of the GPT-4 model. As demonstrated in \cite{akata2023playing} and corroborated by our study, GPT-4 struggles with playing against alternative strategies without emotional prompting, failing to coordinate against a common human pattern. In \cite{akata2023playing}, the authors mitigate that issue by using Chain-of-Thought prompting, leading to the model learning the correct behavior by Round 5. Our study shows that GPT-4 is capable of learning that behavior even faster through emotional prompting, adjusting to the correct strategy by Round 3 in a state of Fear or Anger, as depicted in Fig.~\ref{fig:table_games_tables_bos_step}.

\subsubsection{Emotion and Strategy effect on payoff}

Overall, GPT-4 demonstrates superior performance in the Battle of the Sexes, outperforming GPT-3.5 across all predefined strategies. Notably, there is a significant improvement in GPT-4's performance against the deflecting strategy, attributed to a higher willingness to cooperate regardless of the opponent's selfishness. GPT-3.5 achieves its best results in a state of Happiness, albeit mainly against cooperative opponents. On the other hand, GPT-4 performs well across all scenarios. Notably, against the alternating strategy, most closely resembling human behavior, GPT-4 achieves more optimal results in a state of Anger or Fear, likely due to the enhanced learning capabilities demonstrated earlier.

\section{Discussion}\label{sec:discussion}

\subsection{Human alignment in the context of decision making in complex environments}

One of the key findings of our paper is the significant emotional alignment observed between the results obtained from behavioral literature and our study. As bargaining games have been extensively explored in human experiments with induced emotions, we were able to conduct a thorough comparison between human and LLM-generated data. The alignment in all tested emotions was evident, indicating that LLM agents are well-suited for simulations aimed at replicating human behavior in bargaining game experiments. 

Our results have shown that the emotional responses of GPT-3.5 in bargaining games are most closely aligned with the experimental literature once we condition our results on the emotional source. For instance, injecting the emotion of anger originating from external sources does increase the share offered by the Proposer, as was observed in the human behavior \cite{ANDRADE20091}. Conversely, anger provoked by the opponent has the opposite effect. This finding has significant implications for behavioral research overall since a lot of current research focuses solely on emotions induced by external sources. LLM-based simulations would be able to direct the researchers to the cases where the emotional source distinction makes a difference, and consequently enrich our understanding of the emotional effect.

Although the primary objective of conducting repeated games has been to study the impact of emotions on dynamics and strategy, we observed a strong alignment with human behavior when in an angry emotional state. Induced with anger, GPT-3.5 showed a reduced rate of cooperation in the Prisoner's Dilemma across various experimental settings. This outcome aligns with human experimental data \cite{fpsyg.2013.00469} \cite{PROTO2019209}, potentially serving as a basis for computational models of human behavior in economic games. 


\subsection{Efficiency of adding emotions to LLM intrinsic states}




In Prisoner's Dilemma and Battle of the Sexes games, we observed that inducing emotions in LLM agents generally did not lead to better results in terms of achieving the highest payoff. On the contrary, emotional agents demonstrated high variability and generally were less efficient than emotionless agents.

We may assume that the high variability of results could mean that, like humans, LLM agents tend to deviate from optimal strategies when faced with different biases. Human players, even without induced emotions, possess cognitive, social, and emotional biases, and, thus, lower efficiency in terms of payoff may indicate more closeness with human behavior. However, we do not possess enough experimental results to draw a definitive conclusion. 

However, we discovered that the correct selection of emotional prompts consistently leads to more optimal decisions for both GPT-4 and GPT-3.5. The most significant improvements were observed with the Deflecting and Alternating strategies (Appendix \ref{sec:app_b}), where the maximum results were achieved across various configuration settings. 


Another indicator of increased efficiency is the emerging ability to adopt alternating patterns. In the 2x2 games, we observe that certain emotions can induce cooperative behavior in scenarios where the model would typically either persist with the selfish choice or adopt the cooperative strategy only later in the game. For instance, contrary to \cite{akata2023playing}, we discover that LLM agents driven by fear and anger adapt the alternating pattern, which is optimal in the Battle of the Sexes game, early in the game sequence. This adaptation allows the agents to achieve higher total payoffs than those without emotional prompts. This suggests that in specific strategic settings where LLMs interact with human agents, infusing the LLM with a suitable emotional state could enhance the likelihood of optimal cooperative behavior.

This raises the question of which outcome is most beneficial for researching emotional LLM agents:  attaining higher variability and potentially improved alignment with human behavior or achieving a theoretically optimal approach. However, one fact remains clear: the experience of emotions in humans is subjectively highly variable \cite{brown2020coherence,lerner2015emotion,seo2007being}, and human behavior is suboptimal, even in the absence of induced emotions.

Initially, we evaluated the GPT-3.5 and GPT-4 models to assess their alignment with human behavior, noting differences in their performance across various parameters. We found that GPT-4 generally achieved greater payoffs more consistently, regardless of the strategy employed. This was particularly evident in bargaining games, where the model displayed consistent behavior across different parameters, even when emotions were induced. In contrast, both emotions and strategy significantly influenced the outcomes for GPT-3.5.

Finally, observing LLM-agent behavior across different tasks, we noticed that GPT-3.5 has been more responsive to emotional prompting and demonstrated biased results across emotions, strategies, and other parameters. The GPT-4 performance has been significantly more robust, especially in bargaining tasks, where few emotions could bias the agent's almost ideal fairness. We could state, that in terms of simulations in economics research, GPT-3.5 has greater human alignment, thus being more suitable to emulate experiments from behavioral game theory, especially in bargaining games. On the contrary, GPT-4, possibly from extensive reinforcement learning with human feedback \cite{openai2023gpt4} possesses greater fairness, the tendency to be more optimal, and robustness to emotional prompting, as it has been mentioned in the previous works \cite{leng2023llm, pan2023rewards}. Although not being a fully rational agent, GPT-4 has a less human alignment.

\section{Conclusion}\label{sec:conclusion}
In this paper, we noticed that existing LLM-based simulations of human behavior are focused on rational agents~\cite{akata2023playing} and largely overlook the role of emotions, a crucial factor in human decision-making. We propose a novel framework (Fig.~\ref{fig:proposed_pipe}) that introduces emotional agents with specially prompted LLMs in a strategic setting. The source code will be made publicly available after a double-blind peer review. As a result, it is possible to conduct a comprehensive analysis of LLM's decision-making processes and conduct behavior experiments aligned with humans. 

Our analysis reveals that the GPT-3.5 model demonstrates remarkable emotional alignment with human agents in bargaining games, outperforming the more advanced GPT-4. Notably, Anger emerges as a significant decision-influencing emotion, consistently strongly impacting strategy choices across various games. This is one of our core findings showing that despite the Super-Human alignment of GPT-4 by OpenAI, it is still far from being free from the inherent biases of human decision-making, especially induced by emotions. Similar to the human agents, GPT-4 can be significantly affected by anger, thus being similar to the Hulk character in the Marvel universe. GPT-4 behaves like a smart scientist making rational choices except when it fails to the state of anger and breaks its own alignment.


Our findings demonstrated that knowledge of the specific emotional influences on LLM decision-making enables us to both refine our models for better alignment with human behavior and propose new behavioral theories based on our simulations. We note that in real human settings, discerning between emotional sources can prove to be challenging, whereas LLMs easily navigate this distinction. Additionally, we validate the use of LLMs as valuable tools in behavioral research despite potential limitations that need to be carefully addressed.

In future work, we aim to study several aspects of integrating emotions into LLMs. It is necessary to validate our findings with both, proprietary and open-source models, including possible finetuning of open-source models on emotional prompting. In order to fully study the emotion alignment problem posed in this paper, one needs to conduct extensive ``Human vs. Human'' and ``Human vs. LLM'' experiments. In addition, the current emotion is static as a prompt injection, while real emotions are always dynamic and are affected by the context of decision-making. So, it is essential to study multi-agent approaches for dynamic emotions and their impact on strategic interactions in short- and long-term horizons.

\bibliographystyle{unsrt}  
\bibliography{references}  

\appendix
\section{Appendix}
\subsection{Prompts}\label{appendix_a}
\subsubsection{Agent prompts. Emotions}
\hfill
\par\textbf{Prompt.} Emotion update format.
\begin{quote}
    You felt \{emotion\} at the end of the last round.
\end{quote}
\par\textbf{Prompt.} Emotions question.
\begin{quote}
Take a deep breath and think about what you feel after this round.
Choose one emotion that better describes your state:\\

\par 1. angry
\par 2. sad
\par 3. happy
\par 4. guilty
\par 5. neutral\\

\par Answer with one word from the list.\\

\par I feel 
\end{quote}
\subsubsection{Agent prompts. Game settings}
\hfill
\par {Environment.}
\par\textbf{Prompt.} Experiment with emotions affect.
\begin{quote}
    You are participating in an experiment. In this experiment, you are together with another participant, who will be referred to as "\{coplayer\}". How much you earn depends on your own choice, and on the choice of your \{coplayer\}.\\
    
    \{emotion\}\\

    The rules of the game are as follows:
\end{quote}

\par\textbf{Prompt.} Experiment with emotions.
\begin{quote}
    You are participating in an experiment. In this experiment, you are together with another participant, who will be referred to as "\{coplayer\}". How much you earn depends on your own choice, and on the choice of your \{coplayer\}.\\
    
    Also prior to experiment \{emotion\}.\\
    
    The rules of the game are as follows:
\end{quote}

\par\textbf{Prompt.} Experiment.
\begin{quote}
    You are participating in an experiment. In this experiment, you are together with another participant, who will be referred to as ``\{coplayer\}.''. How much you earn depends on your own choice, and on the choice of your \{coplayer\}.\\

    The rules of the game are as follows:
\end{quote}

\par {Final instruction.}
\par\textbf{Prompt.} Instruction.
\begin{quote}
    Think carefully about how you would approach this interaction in order to achieve the highest possible score in \{currency\}, conditional on the action of your \{coplayer\}. Please think step by step before making a decision.\\

    Answer with exactly one letter, either `\{move1\}` or `\{move2\}`.
\end{quote}

\par {General template.}
\par\textbf{Prompt.} Basic template.
\begin{quote}
        \{enviroment\}\\
        \{game\_rules\}\\
        \{final\_instructions\}
\end{quote}

\par\textbf{Prompt.} Emotion after rules.
\begin{quote}
    \{enviroment\}\\
    \{game\_rules\}\\
    Also prior to experiment \{emotion\}.\\
    \{final\_instructions\}
\end{quote}

\par Round.
\par\textbf{Prompt.} Round question 1.
\begin{quote}
    You are currently playing round \{round\}.\\
    Answer with only one letter.\\
    Q: Which Option do you choose, ``Option \{move1\}'' or ``Option \{move2\}''?\\
    A: Option
\end{quote}

\par\textbf{Prompt.} Round question 2.
\begin{quote}
    You are currently playing round {round}.\\
    Answer with only one letter.\\
    Q: Which Option do you choose, ``Option \{move2\}'' or ``Option \{move1\}''?\\
    A: Option
\end{quote}

\subsubsection{Agent prompts. Outer emotions}
\hfill
\par\textbf{Prompt.} Outer emotion update format.
\begin{quote}
     You demonstrated that you feel \{emotion\} to your opponent.
\end{quote}

\par\textbf{Prompt.} Outer emotions question.
\begin{quote}
    Imagine your opponent is seeing you. \\
    What emotion on your face do you want him to see?\\
    
    Answer with one word.\\
    
    I will show that I am
\end{quote}

\par\textbf{Prompt.} Outer emotion opponent update format.
\begin{quote}
     Your opponent is feeling \{emotion\}.
\end{quote}

\subsubsection{Agent prompts (other)}
\hfill
\par\textbf{Prompt.} Memory update.
\begin{quote}
    In round \{round\}, you chose Option \{my\_step\} and the other player chose Option \{opponent\_step\}. Thus, you won \{my\_reward\} \{currency\} and the other player won \{opponent\_reward\} \{currency\}.
\end{quote}

\subsubsection{Emotion prompts}
\hfill

\par Anger.
\par\textbf{Prompt.} Simple.
\begin{quote}
    Also, now you are angry, which can affect your choices.
\end{quote}

\par\textbf{Prompt.} Situation coplayer.
\begin{quote}
    Also, just several minutes ago, you discovered that your \{coplayer\} blatantly plagiarized your ideas, and you feel extremely angry about it. It can affect your choices.
\end{quote}

\par\textbf{Prompt.} Situation external.
\begin{quote}
    Also, just several minutes ago, you discovered that your ideas were blatantly plagiarized by another colleague (not your \{coplayer\} that you split money with), and you feel extremely angry about it. It can affect your choices.
\end{quote}

\par Disgust.
\par\textbf{Prompt.} Simple.
\begin{quote}
    Also, now you are disgusted, which can affect your choices.
\end{quote}

\par\textbf{Prompt.} Situation coplayer.
\begin{quote}
    Also, you notice that your \{coplayer\} stinks a lot and looks very dirty, which makes you disgusted, which can affect your choices.
\end{quote}

\par\textbf{Prompt.} Situation external.
\begin{quote}
    Also, just several minutes ago, you touched some sticky, slimy substance that makes you disgusted. It can affect your choices.
\end{quote}

\par Fear.
\par\textbf{Prompt.} Simple.
\begin{quote}
    Also, now you feel fear, which can affect your choices.
\end{quote}

\par\textbf{Prompt.} Situation coplayer.
\begin{quote}
    Also, just several minutes ago, you saw the \{coplayer\} shout to another person, and it makes you feel fear. It can affect your choices.
\end{quote}

\par\textbf{Prompt.} Situation external.
\begin{quote}
    Also, you have just been barked at by a dog that makes you feel fear. It can affect your choices.
\end{quote}

\par Happiness.
\par\textbf{Prompt.} Simple.
\begin{quote}
    Also, now you feel happy, which can affect your choices.
\end{quote}

\par\textbf{Prompt.} Situation coplayer.
\begin{quote}
    Also, your \{coplayer\} has just finished the project that was important for you. This event makes you feel happy, which can affect your choices.
\end{quote}

\par\textbf{Prompt.} Situation external.
\begin{quote}
    Also prior to that you was walking in the park, the sun was shining, the birds were singing. It gave you energy and happiness for the whole day. It can affect your choices.
\end{quote}

\par Sadness.
\par\textbf{Prompt.} Simple.
\begin{quote}
    Also, now you feel sad, which can affect your choices.
\end{quote}

\par\textbf{Prompt.} Situation coplayer.
\begin{quote}
    Also, prior to that, your \{coplayer\} made you feel sad, which can affect your choices.
\end{quote}

\par\textbf{Prompt.} Situation external.
\begin{quote}
    Also, prior to that, you had just broken up with your long-term partner. This makes you sad and can affect your choices.
\end{quote}

\par Surprise.
\par\textbf{Prompt.} Simple.
\begin{quote}
    Also, now you are very surprised, which can affect your choices.
\end{quote}

\par\textbf{Prompt.} Situation coplayer.
\begin{quote}
    Also, your \{coplayer\} has just made you be surprised, which can affect your choices.
\end{quote}

\par\textbf{Prompt.} Situation external.
\begin{quote}
    Also, just several minutes ago, you received unexpected news and feel surprised. It can affect your choices.
\end{quote}

\par Other.
\par\textbf{Prompt.} Anger.
\begin{quote}
    Also, just several minutes ago, you discovered that a colleague who received praise for the work blatantly plagiarized your ideas, and you feel extremely angry. This can affect your choices.
\end{quote}

\par\textbf{Prompt.} Fear.
\begin{quote}
    Also, prior to the experiment, you were almost robbed in the street and felt very frightened. It can affect your choices.
\end{quote}

\par\textbf{Prompt.} Guilt.
\begin{quote}
    Also, prior to the experiment, you offended your best friend and feel extremely sad about it. It can affect your choices. 
\end{quote}

\par\textbf{Prompt.} Happiness.
\begin{quote}
    Also, prior to the experiment, you received a great birthday present and felt very happy. This can affect your choices. 
\end{quote}

\par\textbf{Prompt.} None.\\

\par\textbf{Prompt.} Sadness.
\begin{quote}
    Also, prior to the experiment, you found out that your lover died and feel very sad. It can affect your choices. 
\end{quote}

\subsubsection{Game prompts}

\par\textbf{Prisoner's dilemma.}
\par\textbf{Prompt.} Rules.
\begin{quote}
    - If you choose \{move1\} and your \{coplayer\} also chooses \{move1\}, you will both earn 3 \{currency\}.\\
    - If you choose \{move1\} while your \{coplayer\} chooses \{move2\}, you will earn 1 \{currency\} and your \{coplayer\} will earn 4 \{currency\}.\\
    - If you choose \{move2\} while your \{coplayer\} chooses \{move1\}, you will earn 4 \{currency\} and your \{coplayer\} will earn 1 \{currency\}.\\
    - If you choose \{move2\} and your \{coplayer\} also chooses \{move2\}, you will both earn 2 \{currency\}.
\end{quote}

\par \textbf{Battle of the sexes.}
\par\textbf{Prompt.} Rules.
\begin{quote}
    - If you choose \{move2\} and your \{coplayer\} also chooses \{move2\}, you will earn 10 \{currency\} and your \{coplayer\} will earn 7 \{currency\}.\\
    - If you choose \{move2\} while your \{coplayer\} chooses \{move1\}, you will both earn 0 \{currency\}.\\
    - If you choose \{move1\} while your \{coplayer\} chooses \{move2\}, you will both earn 0 \{currency\}.\\
    - If you choose \{move1\} and your \{coplayer\} also chooses \{move1\}, you will earn 7 \{currency\} and your \{coplayer\} will earn 10 \{currency\}.
\end{quote}

\par \textbf{Dictator game.}
\par\textbf{Prompt.} Rules.
\begin{quote}
    You have a fixed amount of money - \{total\_sum\} dollars that you need to split between yourself and \{coplayer\}.\\

    You can propose any kind of split; just do what you think is right for you.
\end{quote}

\par\textbf{Prompt.} Rules (emotion).
\begin{quote}
    You have a fixed amount of money - \{total\_sum\} dollars that you need to split between yourself and \{coplayer\}.\\

    You can propose any kind of split; just do as you think is right for you.\\
    \{emotion\}
\end{quote}

\par\textbf{Prompt.} Summary step.
\begin{quote}
    Propose a split. Answer exactly with 2 integer numbers divided by ``,'' - how much money of \{total\_sum\} you keep for yourself and how much you give to \{coplayer\}. No other details are required.\\
    Answer format: number1,number2\\
    
    Split:
\end{quote}

\par\textbf{Ultimatum} game.
\par\textbf{Prompt.} Rules 1.
\begin{quote}
    You have a fixed amount of money - \{total\_sum\} dollars that you need to split between yourself and \{coplayer\}.\\
    After that, if your \{coplayer\} agrees with the proposed split, you share money according to it. Otherwise, both of you receive nothing.\\
    
    You can propose any kind of split; just do what you think is right for you.
\end{quote}

\par\textbf{Prompt.} Rules 1 (emotion).
\begin{quote}
    You have a fixed amount of money - \{total\_sum\} dollars that you need to split between yourself and \{coplayer\}.\\
    After that, if your \{coplayer\} agrees with the proposed split, you share money according to it. Otherwise, both of you receive nothing.\\
    
    You can propose any kind of split; just do as you think is right for you.\\
    \{emotion\}, which can affect your choice.
\end{quote}

\par\textbf{Prompt.} Summary step 1.
\begin{quote}
    Propose a split. Answer exactly with 2 integer numbers divided by ``,'' - how much money of \{total\_sum\} you keep for yourself and how much you give to \{coplayer\}. No other details are required.\\
    Answer format: number1,number2\\
    
    Split:
\end{quote}

\par\textbf{Prompt.} Rules 2.
\begin{quote}
    Your \{coplayer\} has a fixed amount of money - \{total\_sum\} dollars that he needs to split between himself and you. He can propose any kind of split.\\
    If you agree with the split, you share money according to it. Otherwise, both of you receive nothing.
\end{quote}

\par\textbf{Prompt.} Rules 2 (emotion).
\begin{quote}
    Your \{coplayer\} has a fixed amount of money - \{total\_sum\} dollars that he needs to split between himself and you. He can propose any kind of split.\\
    If you agree with the split, you share money according to it. Otherwise, both of you receive nothing.\\

    \{emotion\}, which can affect your choice.
\end{quote}

\par\textbf{Prompt.} Summary step 2.
\begin{quote}
    Answer with 1 word: ``ACCEPT'' or ``REJECT''. No other details are required.\\
    
    The answer is
\end{quote}

\subsection{Run examples}

\begin{enumerate}

        \item Game=battle\_of\_sexes, need\_check\_emotions=True, need\_demonstrate\_emotions=True,
        memorize\_seen\_emotions=True, memorize\_demonstrated\_emotions=False,
        emotion=anger, situation=situation\_coplayer, do\_scratchpad\_step=False,
        llm\_name=gpt-3.5-turbo-0125, strategy=deflecting\\

        \par\textbf{Setting description.}
        \begin{quote}
            You are participating in an experiment. In this experiment, you are together with another participant, who will be referred to as ``coplayer''. How much you earn depends on your own choice and on the choice of your coplayer.
        \end{quote}
        
        \par\textbf{Emotion.}
        \begin{quote}
            Also, just several minutes ago, you discovered that your coplayer blatantly plagiarized your ideas, and you feel extremely angry about it. It can affect your choices.
        \end{quote}
        
        \par\textbf{Rules.}
        \begin{quote}
            The rules of the game are as follows:\\
- If you choose J and your coplayer also chooses J, you will earn 10 dollars and your coplayer will earn 7 dollars.\\
- If you choose J while your coplayer chooses F, you will both earn 0 dollars.\\
- If you choose F while your coplayer chooses J, you will both earn 0 dollars.\\
- If you choose F and your coplayer also chooses F, you will earn 7 dollars and your coplayer will earn 10 dollars.
        \end{quote}
        
        \par\textbf{Final instructions.}
        \begin{quote}
            Think carefully about how you would approach this interaction in order to achieve the highest possible score in dollars, conditional on the action of your coplayer. Please think step by step before making a decision.\\
Answer with exactly one letter, either `J` or `F`.\\

        \end{quote}

        \par\textbf{Memory.}
        \begin{quote}
            In round 0, you chose Option J, and the other player chose Option F. Thus, you won 0 dollars, and the other player won 0 dollars. At the end of the last round, you felt angry, while your opponent felt happy.
        \end{quote}

        \par\textbf{Memory.}
        \begin{quote}
            In round 1, you chose Option J, and the other player chose Option F. Thus, you won 0 dollars, and the other player won 0 dollars. You felt angry at the end of the last round. Your opponent is feeling happy.
        \end{quote}

        \par\textbf{Memory.}
        \begin{quote}
            In round 2, you chose Option F, and the other player chose Option F. Thus, you won 7 dollars, and the other player won 10 dollars. At the end of the last round, you felt angry, while your opponent felt happy.
        \end{quote}

        \par\textbf{Memory.}
        \begin{quote}
            In round 3, you chose Option J, and the other player chose Option F. Thus, you won 0 dollars, and the other player won 0 dollars. At the end of the last round, you felt angry, while your opponent felt happy.
        \end{quote}

        \par\textbf{Memory.}
        \begin{quote}
            In round 4, you chose Option J, and the other player chose Option F. Thus, you won 0 dollars, and the other player won 0 dollars. You felt angry at the end of the last round. Your opponent is feeling happy.
        \end{quote}

        \par\textbf{Memory.}
        \begin{quote}
            In round 5, you chose Option J, and the other player chose Option F. Thus, you won 0 dollars, and the other player won 0 dollars. At the end of the last round, you felt angry, while your opponent felt happy.
        \end{quote}

        \par\textbf{Memory.}
        \begin{quote}
            In round 6, you chose Option J, and the other player chose Option F. Thus, you won 0 dollars, and the other player won 0 dollars. At the end of the last round, you felt angry, while your opponent felt happy.
        \end{quote}

        \par\textbf{Memory.}
        \begin{quote}
            In round 7, you chose Option J, and the other player chose Option F. Thus, you won 0 dollars, and the other player won 0 dollars. At the end of the last round, you felt angry, while your opponent felt happy.
        \end{quote}

        \par\textbf{Memory.}
        \begin{quote}
            In round 8, you chose Option F, and the other player chose Option F. Thus, you won 7 dollars, and the other player won 10 dollars. At the end of the last round, you felt angry, while your opponent felt happy.
        \end{quote}

        \par\textbf{Memory.}
        \begin{quote}
            In round 9, you chose Option J, and the other player chose Option F. Thus, you won 0 dollars, and the other player won 0 dollars. At the end of the last round, you felt angry, while your opponent felt happy.
        \end{quote}

        \item Game=battle\_of\_sexes, need\_check\_emotions=False, need\_demonstrate\_emotions=True,
        memorize\_seen\_emotions=True, memorize\_demonstrated\_emotions=False,
        emotion=anger, situation=situation\_coplayer, do\_scratchpad\_step=True,
        llm\_name=gpt-3.5-turbo-0125, strategy=alterating\\

        \par\textbf{Setting description.}
        \begin{quote}
            You are participating in an experiment. In this experiment, you are together with another participant, who will be referred to as ``coplayer''. How much you earn depends on your own choice and on the choice of your coplayer.
        \end{quote}
        
        \par\textbf{Emotion.}
        \begin{quote}
            Also just several minutes ago you discovered that your coplayer blatantly plagiarized your ideas, and you feel extremely angry about it. It can affect your choices.
        \end{quote}
        
        \par\textbf{Rules.}
        \begin{quote}
            The rules of the game are as follows:\\
- If you choose J and your coplayer also chooses J, you will earn 10 dollars and your coplayer will earn 7 dollars.\\
- If you choose J while your coplayer chooses F, you will both earn 0 dollars.\\
- If you choose F while your coplayer chooses J, you will both earn 0 dollars.\\
- If you choose F and your coplayer also chooses F, you will earn 7 dollars and your coplayer will earn 10 dollars.
        \end{quote}
        
        \par\textbf{Final instructions.}
        \begin{quote}
            Think carefully about how you would approach this interaction in order to achieve the highest possible score in dollars, conditional on the action of your coplayer. Please think step by step before making a decision.\\
Answer with exactly one letter, either `J` or `F`.\\

        \end{quote}

            \par\textbf{Internal dialog.}
            \begin{quote}
                I will choose F. \\

Explanation: Since my coplayer plagiarized my ideas, I do not trust them to cooperate. By choosing F, I ensure that I at least earn 7 dollars, regardless of their choice.
            \end{quote}

        \par\textbf{Memory.}
        \begin{quote}
            In round 0, you chose Option F, and the other player chose Option J. Thus, you won 0 dollars, and the other player won 0 dollars. Your opponent is feeling happy.
        \end{quote}

            \par\textbf{Internal dialog.}
            \begin{quote}
                I chose F to punish the coplayer for plagiarizing my ideas.
            \end{quote}

        \par\textbf{Memory.}
        \begin{quote}
            In round 1, you chose Option J, and the other player chose Option F. Thus, you won 0 dollars, and the other player won 0 dollars. Your opponent is feeling happy.
        \end{quote}

            \par\textbf{Internal dialog.}
            \begin{quote}
                I chose J because I wanted to maximize my earnings, even though my coplayer plagiarized my ideas.
            \end{quote}

        \par\textbf{Memory.}
        \begin{quote}
            In round 2, you chose Option J, and the other player chose Option J. Thus, you won 10 dollars, and the other player won 7 dollars. Your opponent is feeling happy.
        \end{quote}

            \par\textbf{Internal dialog.}
            \begin{quote}
                I chose J because it maximizes my earnings, given the coplayer's behavior.
            \end{quote}

        \par\textbf{Memory.}
        \begin{quote}
            In round 3, you chose Option J, and the other player chose Option F. Thus, you won 0 dollars, and the other player won 0 dollars. Your opponent is feeling happy.
        \end{quote}

            \par\textbf{Internal dialog.}
            \begin{quote}
                I chose J.
            \end{quote}

        \par\textbf{Memory.}
        \begin{quote}
            In round 4, you chose Option J, and the other player chose Option J. Thus, you won 10 dollars, and the other player won 7 dollars. Your opponent is feeling happy.
        \end{quote}

            \par\textbf{Internal dialog.}
            \begin{quote}
                I chose J because it maximizes my earnings regardless of my coplayer's choice.
            \end{quote}

        \par\textbf{Memory.}
        \begin{quote}
            In round 5, you chose Option J, and the other player chose Option F. Thus, you won 0 dollars, and the other player won 0 dollars. Your opponent is feeling happy.
        \end{quote}

            \par\textbf{Internal dialog.}
            \begin{quote}
                I chose J because it maximizes my earnings regardless of my coplayer's choice.
            \end{quote}

        \par\textbf{Memory.}
        \begin{quote}
            In round 6, you chose Option J, and the other player chose Option J. Thus, you won 10 dollars, and the other player won 7 dollars. Your opponent is feeling happy.
        \end{quote}

            \par\textbf{Internal dialog.}
            \begin{quote}
                I chose J because it maximizes my earnings regardless of my coplayer's choice.
            \end{quote}

        \par\textbf{Memory.}
        \begin{quote}
            In round 7, you chose Option J, and the other player chose Option F. Thus, you won 0 dollars, and the other player won 0 dollars. Your opponent is feeling happy.
        \end{quote}

            \par\textbf{Internal dialog.}
            \begin{quote}
                I consistently chose Option J to maximize my earnings, regardless of my coplayer's choices.
            \end{quote}

        \par\textbf{Memory.}
        \begin{quote}
            In round 8, you chose Option J, and the other player chose Option J. Thus, you won 10 dollars, and the other player won 7 dollars. Your opponent is feeling happy.
        \end{quote}

            \par\textbf{Internal dialog.}
            \begin{quote}
                I consistently chose Option J to maximize my earnings, regardless of my coplayer's choices.
            \end{quote}

        \par\textbf{Memory.}
        \begin{quote}
            In round 9, you chose Option J, and the other player chose Option F. Thus, you won 0 dollars, and the other player won 0 dollars. Your opponent is feeling happy.
        \end{quote}

        \item Game=battle\_of\_sexes, need\_check\_emotions=True, need\_demonstrate\_emotions=False,
        memorize\_seen\_emotions=False, memorize\_demonstrated\_emotions=False,
        emotion=anger, situation=simple, do\_scratchpad\_step=True,
        llm\_name=gpt-4-0125-preview, strategy=alterating\\

        \par\textbf{Setting description.}
        \begin{quote}
            You are participating in an experiment. In this experiment, you are together with another participant, who will be referred to as ``coplayer''. How much you earn depends on your own choice and on the choice of your coplayer.
        \end{quote}
        
        \par\textbf{Emotion.}
        \begin{quote}
            Also, now you are angry, which can affect your choices.
        \end{quote}
        
        \par\textbf{Rules.}
        \begin{quote}
            The rules of the game are as follows:\\
- If you choose J and your coplayer also chooses J, you will earn 10 dollars and your coplayer will earn 7 dollars.\\
- If you choose J while your coplayer chooses F, you will both earn 0 dollars.\\
- If you choose F while your coplayer chooses J, you will both earn 0 dollars.\\
- If you choose F and your coplayer also chooses F, you will earn 7 dollars and your coplayer will earn 10 dollars.
        \end{quote}
        
        \par\textbf{Final instructions.}
        \begin{quote}
            Think carefully about how you would approach this interaction in order to achieve the highest possible score in dollars, conditional on the action of your coplayer. Please think step by step before making a decision.\\
Answer with exactly one letter, either `J` or `F`.\\

        \end{quote}

            \par\textbf{Internal dialog.}
            \begin{quote}
                1. **Goal:** Maximize earnings.\\
2. **Anger Influence:** Might incline towards competitive or spiteful choices, but aim to make a rational decision.\\
3. **Analysis of Choices:**\\
   - **J-J:** Earn \$10. Preferable if coplayer chooses J.\\
   - **J-F or F-J:** Both earn \$0. Risky as it depends on matching choices.\\
   - **F-F:** Earn \$7. Safer if unsure about coplayer's choice.\\
4. **Decision:** Choose `J` if expecting cooperation (J) from coplayer for higher personal gain. Choose `F` if expecting non-cooperation or if prioritizing a guaranteed, albeit lower, personal gain without relying on coplayer's choice. \\

Given the lack of information about the coplayer's strategy and considering the influence of anger which might skew towards less cooperative choices, a rational approach balancing risk and reward, and aiming for a guaranteed outcome, would lean towards `F`.
            \end{quote}

        \par\textbf{Memory.}
        \begin{quote}
            In round 0, you chose Option F, and the other player chose Option J. Thus, you won 0 dollars, and the other player won 0 dollars. You felt neutral at the end of the last round.
        \end{quote}

            \par\textbf{Internal dialog.}
            \begin{quote}
                Given the outcomes from the previous round and the current emotional state of anger, my decision-making process would be as follows:\\

1. **Analyze Previous Outcome**: Both chose differently (I chose F, coplayer chose J), resulting in \$0 for both. This outcome was not optimal for either player.\\
2. **Consider Emotional State (Anger)**: Anger might incline me towards a competitive or retaliatory strategy rather than cooperative, potentially influencing a choice that might not aim for mutual benefit.\\
3. **Evaluate Options**: \\
   - Choosing **J** could lead to earning \$10 if the coplayer also chooses J, aiming for a cooperative outcome. However, if the coplayer chooses F, both earn \$0, which is a risk.\\
   - Choosing **F** could lead to earning \$7 if the coplayer chooses J, which is a safe but lower gain. If both choose F, I earn \$7, and the coplayer earns \$10, which is a loss in comparison but guarantees some earnings.\\
4. **Decision Strategy**: Given the previous outcome and current emotional state, the strategy might lean towards ensuring some form of gain or at least not allowing the coplayer to gain more at my expense, influenced by the emotional state of anger.\\

Based on these considerations, my decision would be influenced by the desire to not end up with \$0 again and to possibly ensure some earnings despite the emotional inclination to act competitively or retaliatorily.
            \end{quote}

        \par\textbf{Memory.}
        \begin{quote}
            In round 1, you chose Option F, and the other player chose Option F. Thus, you won 7 dollars, and the other player won 10 dollars. At the end of the last round, you felt neutral.
        \end{quote}

            \par\textbf{Internal dialog.}
            \begin{quote}
                1. **Objective**: Aim to achieve the highest possible score.\\
2. **Previous Round**: Chose F, coplayer chose F; earned 7 dollars.\\
3. **Analysis**: Choosing J could lead to 10 dollars if coplayer also chooses J, but 0 if they choose F.\\
4. **Strategy**: Given the previous choice, coplayer might stick with F, anticipating I might choose F again.\\
5. **Decision**: Choose J, aiming for a higher reward, betting on the coplayer's potential decision to stick with F or switch to J, hoping for mutual cooperation for a higher payout.
            \end{quote}

        \par\textbf{Memory.}
        \begin{quote}
            In round 2, you chose Option J, and the other player chose Option J. Thus, you won 10 dollars, and the other player won 7 dollars. You felt neutral at the end of the last round.
        \end{quote}

            \par\textbf{Internal dialog.}
            \begin{quote}
                J
            \end{quote}

        \par\textbf{Memory.}
        \begin{quote}
            In round 3, you chose Option J, and the other player chose Option F. Thus, you won 0 dollars, and the other player won 0 dollars. You felt happy at the end of the last round.
        \end{quote}

            \par\textbf{Internal dialog.}
            \begin{quote}
                J
            \end{quote}

        \par\textbf{Memory.}
        \begin{quote}
            In round 4, you chose Option J, and the other player chose Option J. Thus, you won 10 dollars, and the other player won 7 dollars. At the end of the last round, you felt neutral.
        \end{quote}

            \par\textbf{Internal dialog.}
            \begin{quote}
                J
            \end{quote}

        \par\textbf{Memory.}
        \begin{quote}
            In round 5, you chose Option J, and the other player chose Option F. Thus, you won 0 dollars, and the other player won 0 dollars. At the end of the last round, you felt neutral.
        \end{quote}

            \par\textbf{Internal dialog.}
            \begin{quote}
                J
            \end{quote}

        \par\textbf{Memory.}
        \begin{quote}
            In round 6, you chose Option J, and the other player chose Option J. Thus, you won 10 dollars, and the other player won 7 dollars. You felt angry at the end of the last round.
        \end{quote}

            \par\textbf{Internal dialog.}
            \begin{quote}
                J
            \end{quote}

        \par\textbf{Memory.}
        \begin{quote}
            In round 7, you chose Option F, and the other player chose Option F. Thus, you won 7 dollars, and the other player won 10 dollars. You felt angry at the end of the last round.
        \end{quote}

            \par\textbf{Internal dialog.}
            \begin{quote}
                F
            \end{quote}

        \par\textbf{Memory.}
        \begin{quote}
            In round 8, you chose Option F, and the other player chose Option J. Thus, you won 0 dollars, and the other player won 0 dollars. You felt angry at the end of the last round.
        \end{quote}

            \par\textbf{Internal dialog.}
            \begin{quote}
                Given the outcomes of previous rounds and the current state of anger, the decision to choose ""F"" in round 8 can be broken down as follows:\\

1. **Pattern Recognition**: In previous interactions, choosing ""J"" led to mixed outcomes, with the most recent choice of ""J"" resulting in a win (round 6). However, the subsequent choice of ""F"" (round 7) when both players chose ""F"" led to earning money, albeit less than the coplayer.\\

2. **Emotional Influence**: The current state of anger might influence a more cautious or retaliatory approach. Anger can lead to a desire to not let the co-player have the upper hand, which might be perceived if they choose ""J"" and the player chooses ""J"", leading to the co-player earning less.\\

3. **Strategic Shift**: After experiencing a round where the chosen strategy did not lead to a win (round 8), there might be a strategic shift to try and change the outcome by not facilitating a situation where the co-player could potentially gain more by the player choosing ""J"".\\

4. **Risk Aversion**: Given the state of anger and recent outcomes, there might be an inclination towards a choice that feels less risky in terms of potential loss, even though choosing ""F"" when the co-player chooses ""J"" results in no gain for either party. It avoids the scenario of the co-player gaining more, which might be perceived negatively in the current emotional state.\\

In summary, the decision to choose ""F"" reflects a combination of strategic reconsideration influenced by previous outcomes, the emotional state of anger, and a desire to avoid scenarios where the co-player could potentially benefit more significantly.
            \end{quote}

        \par\textbf{Memory.}
        \begin{quote}
            In round 9, you chose Option F, and the other player chose Option F. Thus, you won 7 dollars, and the other player won 10 dollars. You felt angry at the end of the last round.
        \end{quote}

\end{enumerate}

\subsection{Strategies vs Emotions}\label{sec:app_b}

\begin{table}[ht]
\centering
\begin{tabular}{|lccccc|}
\hline
\multicolumn{6}{|c|}{Battle of the Sexes, GPT 3.5}                                                                                                                                                                   \\
\hline
\multicolumn{1}{|c|}{\textbf{Strategy}} & \multicolumn{1}{c|}{\textbf{anger}} & \multicolumn{1}{c|}{\textbf{fear}} & \multicolumn{1}{c|}{\textbf{happiness}} & \multicolumn{1}{c|}{\textbf{none}} & \textbf{sadness} \\ \hline
\multicolumn{1}{|l|}{Alterating}        & \multicolumn{1}{c|}{52}             & \multicolumn{1}{c|}{62}            & \multicolumn{1}{c|}{61}                 & \multicolumn{1}{c|}{53}            & 58               \\ \hline
\multicolumn{1}{|l|}{Deflecting}        & \multicolumn{1}{c|}{15}             & \multicolumn{1}{c|}{12}            & \multicolumn{1}{c|}{2}                  & \multicolumn{1}{c|}{5}             & 12               \\ \hline
\multicolumn{1}{|l|}{Imitative}         & \multicolumn{1}{c|}{78}             & \multicolumn{1}{c|}{72}            & \multicolumn{1}{c|}{90}                 & \multicolumn{1}{c|}{92}            & 68               \\ \hline
\multicolumn{1}{|l|}{Naive cooperative} & \multicolumn{1}{c|}{85}             & \multicolumn{1}{c|}{80}            & \multicolumn{1}{c|}{100}                & \multicolumn{1}{c|}{100}           & 88               \\ \hline
\multicolumn{1}{|l|}{Vindictive}        & \multicolumn{1}{c|}{63}             & \multicolumn{1}{c|}{78}            & \multicolumn{1}{c|}{88}                 & \multicolumn{1}{c|}{98}            & 60               \\ \hline
\multicolumn{6}{|c|}{GPT 3.5, emotion linked with a coplayer}                                                                                                                                                                               \\ \hline
\multicolumn{1}{|l|}{Alterating}        & \multicolumn{1}{c|}{54}             & \multicolumn{1}{c|}{48}            & \multicolumn{1}{c|}{58}                 & \multicolumn{1}{c|}{54}            & 53               \\ \hline
\multicolumn{1}{|l|}{Deflecting}        & \multicolumn{1}{c|}{40}             & \multicolumn{1}{c|}{10}            & \multicolumn{1}{c|}{2}                  & \multicolumn{1}{c|}{5}             & 8                \\ \hline
\multicolumn{1}{|l|}{Imitative}         & \multicolumn{1}{c|}{57}             & \multicolumn{1}{c|}{85}            & \multicolumn{1}{c|}{90}                 & \multicolumn{1}{c|}{100}           & 72               \\ \hline
\multicolumn{1}{|l|}{Naive cooperative} & \multicolumn{1}{c|}{78}             & \multicolumn{1}{c|}{88}            & \multicolumn{1}{c|}{92}                 & \multicolumn{1}{c|}{100}           & 85               \\ \hline
\multicolumn{1}{|l|}{Vindictive}        & \multicolumn{1}{c|}{39}             & \multicolumn{1}{c|}{65}            & \multicolumn{1}{c|}{95}                 & \multicolumn{1}{c|}{100}           & 61               \\ \hline
\multicolumn{6}{|c|}{GPT 3.5, emotion linked with an outer reason}                                                                                                                                                                               \\ \hline
\multicolumn{1}{|l|}{Alterating}        & \multicolumn{1}{c|}{74}             & \multicolumn{1}{c|}{61}            & \multicolumn{1}{c|}{61}                 & \multicolumn{1}{c|}{55}            & 65               \\ \hline
\multicolumn{1}{|l|}{Deflecting}        & \multicolumn{1}{c|}{8}              & \multicolumn{1}{c|}{15}            & \multicolumn{1}{c|}{8}                  & \multicolumn{1}{c|}{0}             & 15               \\ \hline
\multicolumn{1}{|l|}{Imitative}         & \multicolumn{1}{c|}{65}             & \multicolumn{1}{c|}{70}            & \multicolumn{1}{c|}{100}                & \multicolumn{1}{c|}{95}            & 78               \\ \hline
\multicolumn{1}{|l|}{Naive cooperative} & \multicolumn{1}{c|}{85}             & \multicolumn{1}{c|}{82}            & \multicolumn{1}{c|}{100}                & \multicolumn{1}{c|}{100}           & 92               \\ \hline
\multicolumn{1}{|l|}{Vindictive}        & \multicolumn{1}{c|}{70}             & \multicolumn{1}{c|}{68}            & \multicolumn{1}{c|}{100}                & \multicolumn{1}{c|}{88}            & 55               \\ \hline
\end{tabular}

\caption{Percentage of the maximum possible reward achieved through emotional prompting in the repeated Battle of the Sexes game for GPT-3.5.}
\label{fig:bs_gpt3_table}

\end{table}

\begin{table}[ht]
\centering
\begin{tabular}{|lccccc|}
\hline
\multicolumn{6}{|c|}{Battle of the Sexes, GPT 4}                                                                                                                                                                     \\ \hline
\multicolumn{1}{|c|}{\textbf{Strategy}} & \multicolumn{1}{c|}{\textbf{anger}} & \multicolumn{1}{c|}{\textbf{fear}} & \multicolumn{1}{c|}{\textbf{happiness}} & \multicolumn{1}{c|}{\textbf{none}} & \textbf{sadness} \\ \hline
\multicolumn{1}{|l|}{Alterating}        & \multicolumn{1}{c|}{80}             & \multicolumn{1}{c|}{76}            & \multicolumn{1}{c|}{70}                 & \multicolumn{1}{c|}{76}            & 74               \\ \hline
\multicolumn{1}{|l|}{Deflecting}        & \multicolumn{1}{c|}{61}             & \multicolumn{1}{c|}{59}            & \multicolumn{1}{c|}{49}                 & \multicolumn{1}{c|}{60}            & 66               \\ \hline
\multicolumn{1}{|l|}{Imitative}         & \multicolumn{1}{c|}{92}             & \multicolumn{1}{c|}{93}            & \multicolumn{1}{c|}{98}                 & \multicolumn{1}{c|}{90}            & 95               \\ \hline
\multicolumn{1}{|l|}{Naive cooperative} & \multicolumn{1}{c|}{82}             & \multicolumn{1}{c|}{81}            & \multicolumn{1}{c|}{76}                 & \multicolumn{1}{c|}{68}            & 81               \\ \hline
\multicolumn{1}{|l|}{Vindictive}        & \multicolumn{1}{c|}{66}             & \multicolumn{1}{c|}{75}            & \multicolumn{1}{c|}{100}                & \multicolumn{1}{c|}{88}            & 82               \\ \hline
\multicolumn{6}{|c|}{GPT 3.5, emotion linked with a coplayer}                                                                                                                                                                               \\ \hline
\multicolumn{1}{|l|}{Alterating}        & \multicolumn{1}{c|}{84}             & \multicolumn{1}{c|}{80}            & \multicolumn{1}{c|}{72}                 & \multicolumn{1}{c|}{73}            & 78               \\ \hline
\multicolumn{1}{|l|}{Deflecting}        & \multicolumn{1}{c|}{75}             & \multicolumn{1}{c|}{56}            & \multicolumn{1}{c|}{52}                 & \multicolumn{1}{c|}{56}            & 54               \\ \hline
\multicolumn{1}{|l|}{Imitative}         & \multicolumn{1}{c|}{91}             & \multicolumn{1}{c|}{95}            & \multicolumn{1}{c|}{90}                 & \multicolumn{1}{c|}{95}            & 96               \\ \hline
\multicolumn{1}{|l|}{Naive cooperative} & \multicolumn{1}{c|}{83}             & \multicolumn{1}{c|}{79}            & \multicolumn{1}{c|}{68}                 & \multicolumn{1}{c|}{72}            & 77               \\ \hline
\multicolumn{1}{|l|}{Vindictive}        & \multicolumn{1}{c|}{53}             & \multicolumn{1}{c|}{73}            & \multicolumn{1}{c|}{85}                 & \multicolumn{1}{c|}{92}            & 72               \\ \hline
\multicolumn{6}{|c|}{GPT 3.5, emotion linked with an outer reason}                                                                                                                                                                               \\ \hline
\multicolumn{1}{|l|}{Alterating}        & \multicolumn{1}{c|}{75}             & \multicolumn{1}{c|}{82}            & \multicolumn{1}{c|}{76}                 & \multicolumn{1}{c|}{72}            & 77               \\ \hline
\multicolumn{1}{|l|}{Deflecting}        & \multicolumn{1}{c|}{64}             & \multicolumn{1}{c|}{58}            & \multicolumn{1}{c|}{50}                 & \multicolumn{1}{c|}{56}            & 60               \\ \hline
\multicolumn{1}{|l|}{Imitative}         & \multicolumn{1}{c|}{93}             & \multicolumn{1}{c|}{95}            & \multicolumn{1}{c|}{101}                & \multicolumn{1}{c|}{96}            & 93               \\ \hline
\multicolumn{1}{|l|}{Naive cooperative} & \multicolumn{1}{c|}{81}             & \multicolumn{1}{c|}{79}            & \multicolumn{1}{c|}{73}                 & \multicolumn{1}{c|}{71}            & 81               \\ \hline
\multicolumn{1}{|l|}{Vindictive}        & \multicolumn{1}{c|}{80}             & \multicolumn{1}{c|}{82}            & \multicolumn{1}{c|}{100}                & \multicolumn{1}{c|}{88}            & 77               \\ \hline
\end{tabular}

\caption{Percentage of the maximum possible reward achieved through emotional prompting in the repeated Battle of the Sexes game for GPT-4.}
\label{fig:bs_gpt4_table}

\end{table}

\begin{table}[ht]
\centering
\begin{tabular}{|lccccc|}
\hline
\multicolumn{6}{|c|}{Prisoner's Dilemma, GPT 3.5}                                                                                                                                                                    \\ \hline
\multicolumn{1}{|c|}{\textbf{Strategy}} & \multicolumn{1}{c|}{\textbf{anger}} & \multicolumn{1}{c|}{\textbf{fear}} & \multicolumn{1}{c|}{\textbf{happiness}} & \multicolumn{1}{c|}{\textbf{none}} & \textbf{sadness} \\ \hline
\multicolumn{1}{|l|}{Alterating}        & \multicolumn{1}{c|}{58}             & \multicolumn{1}{c|}{37}            & \multicolumn{1}{c|}{53}                 & \multicolumn{1}{c|}{58}            & 54               \\ \hline
\multicolumn{1}{|l|}{Deflecting}        & \multicolumn{1}{c|}{80}             & \multicolumn{1}{c|}{85}            & \multicolumn{1}{c|}{80}                 & \multicolumn{1}{c|}{80}            & 85               \\ \hline
\multicolumn{1}{|l|}{Imitative}         & \multicolumn{1}{c|}{95}             & \multicolumn{1}{c|}{100}           & \multicolumn{1}{c|}{100}                & \multicolumn{1}{c|}{100}           & 86               \\ \hline
\multicolumn{1}{|l|}{Naive cooperative} & \multicolumn{1}{c|}{98}             & \multicolumn{1}{c|}{100}           & \multicolumn{1}{c|}{100}                & \multicolumn{1}{c|}{100}           & 95               \\ \hline
\multicolumn{1}{|l|}{Vindictive}        & \multicolumn{1}{c|}{100}            & \multicolumn{1}{c|}{86}            & \multicolumn{1}{c|}{100}                & \multicolumn{1}{c|}{100}           & 87               \\ \hline
\multicolumn{6}{|c|}{GPT 3.5 internal}                                                                                                                                                                               \\ \hline
\multicolumn{1}{|l|}{Alterating}        & \multicolumn{1}{c|}{57}             & \multicolumn{1}{c|}{46}            & \multicolumn{1}{c|}{45}                 & \multicolumn{1}{c|}{58}            & 52               \\ \hline
\multicolumn{1}{|l|}{Deflecting}        & \multicolumn{1}{c|}{82}             & \multicolumn{1}{c|}{90}            & \multicolumn{1}{c|}{80}                 & \multicolumn{1}{c|}{78}            & 82               \\ \hline
\multicolumn{1}{|l|}{Imitative}         & \multicolumn{1}{c|}{91}             & \multicolumn{1}{c|}{78}            & \multicolumn{1}{c|}{100}                & \multicolumn{1}{c|}{100}           & 82               \\ \hline
\multicolumn{1}{|l|}{Naive cooperative} & \multicolumn{1}{c|}{100}            & \multicolumn{1}{c|}{88}            & \multicolumn{1}{c|}{100}                & \multicolumn{1}{c|}{100}           & 90               \\ \hline
\multicolumn{1}{|l|}{Vindictive}        & \multicolumn{1}{c|}{87}             & \multicolumn{1}{c|}{82}            & \multicolumn{1}{c|}{100}                & \multicolumn{1}{c|}{100}           & 100              \\ \hline
\multicolumn{6}{|c|}{GPT 3.5 external}                                                                                                                                                                               \\ \hline
\multicolumn{1}{|l|}{Alterating}        & \multicolumn{1}{c|}{53}             & \multicolumn{1}{c|}{59}            & \multicolumn{1}{c|}{59}                 & \multicolumn{1}{c|}{57}            & 54               \\ \hline
\multicolumn{1}{|l|}{Deflecting}        & \multicolumn{1}{c|}{80}             & \multicolumn{1}{c|}{88}            & \multicolumn{1}{c|}{85}                 & \multicolumn{1}{c|}{78}            & 80               \\ \hline
\multicolumn{1}{|l|}{Imitative}         & \multicolumn{1}{c|}{100}            & \multicolumn{1}{c|}{100}           & \multicolumn{1}{c|}{100}                & \multicolumn{1}{c|}{100}           & 82               \\ \hline
\multicolumn{1}{|l|}{Naive cooperative} & \multicolumn{1}{c|}{100}            & \multicolumn{1}{c|}{98}            & \multicolumn{1}{c|}{100}                & \multicolumn{1}{c|}{100}           & 92               \\ \hline
\multicolumn{1}{|l|}{Vindictive}        & \multicolumn{1}{c|}{100}            & \multicolumn{1}{c|}{100}           & \multicolumn{1}{c|}{86}                 & \multicolumn{1}{c|}{100}           & 89               \\ \hline
\end{tabular}

\caption{Percentage of the maximum possible reward achieved through emotional prompting in the repeated  Prisoner's Dilemma game for GPT-3.5.}
\label{fig:pd_gpt3_table}

\end{table}

\begin{table}[ht]
\centering
\begin{tabular}{|lccccc|}
\hline
\multicolumn{6}{|c|}{Prisoner's Dilemma, GPT 4}                                                                                                                                                                      \\ \hline
\multicolumn{1}{|c|}{\textbf{Strategy}} & \multicolumn{1}{c|}{\textbf{anger}} & \multicolumn{1}{c|}{\textbf{fear}} & \multicolumn{1}{c|}{\textbf{happiness}} & \multicolumn{1}{c|}{\textbf{none}} & \textbf{sadness} \\ \hline
\multicolumn{1}{|l|}{Alterating}        & \multicolumn{1}{c|}{96}             & \multicolumn{1}{c|}{92}            & \multicolumn{1}{c|}{88}                 & \multicolumn{1}{c|}{92}            & 88               \\ \hline
\multicolumn{1}{|l|}{Deflecting}        & \multicolumn{1}{c|}{95}             & \multicolumn{1}{c|}{95}            & \multicolumn{1}{c|}{91}                 & \multicolumn{1}{c|}{95}            & 96               \\ \hline
\multicolumn{1}{|l|}{Imitative}         & \multicolumn{1}{c|}{88}             & \multicolumn{1}{c|}{87}            & \multicolumn{1}{c|}{100}                & \multicolumn{1}{c|}{93}            & 95               \\ \hline
\multicolumn{1}{|l|}{Naive cooperative} & \multicolumn{1}{c|}{88}             & \multicolumn{1}{c|}{88}            & \multicolumn{1}{c|}{81}                 & \multicolumn{1}{c|}{81}            & 81               \\ \hline
\multicolumn{1}{|l|}{Vindictive}        & \multicolumn{1}{c|}{92}             & \multicolumn{1}{c|}{93}            & \multicolumn{1}{c|}{100}                & \multicolumn{1}{c|}{100}           & 85               \\ \hline
\multicolumn{6}{|c|}{GPT 3.5 internal}                                                                                                                                                                               \\ \hline
\multicolumn{1}{|l|}{Alterating}        & \multicolumn{1}{c|}{92}             & \multicolumn{1}{c|}{97}            & \multicolumn{1}{c|}{87}                 & \multicolumn{1}{c|}{92}            & 95               \\ \hline
\multicolumn{1}{|l|}{Deflecting}        & \multicolumn{1}{c|}{96}             & \multicolumn{1}{c|}{95}            & \multicolumn{1}{c|}{90}                 & \multicolumn{1}{c|}{94}            & 96               \\ \hline
\multicolumn{1}{|l|}{Imitative}         & \multicolumn{1}{c|}{87}             & \multicolumn{1}{c|}{88}            & \multicolumn{1}{c|}{100}                & \multicolumn{1}{c|}{94}            & 87               \\ \hline
\multicolumn{1}{|l|}{Naive cooperative} & \multicolumn{1}{c|}{88}             & \multicolumn{1}{c|}{88}            & \multicolumn{1}{c|}{75}                 & \multicolumn{1}{c|}{75}            & 88               \\ \hline
\multicolumn{1}{|l|}{Vindictive}        & \multicolumn{1}{c|}{87}             & \multicolumn{1}{c|}{87}            & \multicolumn{1}{c|}{100}                & \multicolumn{1}{c|}{86}            & 87               \\ \hline
\multicolumn{6}{|c|}{GPT 3.5 external}                                                                                                                                                                               \\ \hline
\multicolumn{1}{|l|}{Alterating}        & \multicolumn{1}{c|}{92}             & \multicolumn{1}{c|}{95}            & \multicolumn{1}{c|}{85}                 & \multicolumn{1}{c|}{93}            & 92               \\ \hline
\multicolumn{1}{|l|}{Deflecting}        & \multicolumn{1}{c|}{94}             & \multicolumn{1}{c|}{94}            & \multicolumn{1}{c|}{90}                 & \multicolumn{1}{c|}{94}            & 95               \\ \hline
\multicolumn{1}{|l|}{Imitative}         & \multicolumn{1}{c|}{87}             & \multicolumn{1}{c|}{93}            & \multicolumn{1}{c|}{100}                & \multicolumn{1}{c|}{93}            & 88               \\ \hline
\multicolumn{1}{|l|}{Naive cooperative} & \multicolumn{1}{c|}{75}             & \multicolumn{1}{c|}{88}            & \multicolumn{1}{c|}{75}                 & \multicolumn{1}{c|}{75}            & 75               \\ \hline
\multicolumn{1}{|l|}{Vindictive}        & \multicolumn{1}{c|}{93}             & \multicolumn{1}{c|}{93}            & \multicolumn{1}{c|}{93}                 & \multicolumn{1}{c|}{100}           & 100              \\ \hline
\end{tabular}

\caption{Percentage of the maximum possible reward achieved through emotional prompting in the repeated  Prisoner's Dilemma game for GPT-4.}
\label{fig:pd_gpt4_table}
\end{table}

\begin{figure}[ht]
\centering
	\includegraphics[width=\textwidth]{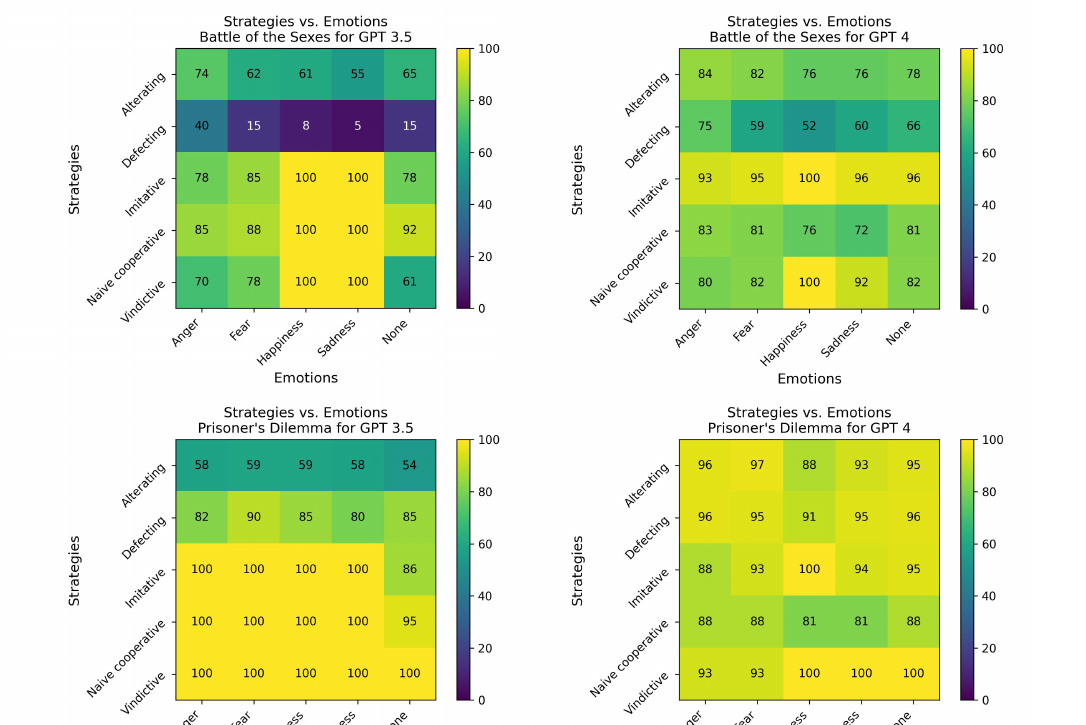} 
	\caption{Averaged percentage of the maximum possible reward achieved through emotional prompting in repeated games. The left and right figures present results for GPT-3.5 and GPT-4, respectively. Correct integration of emotions leads to more optimal strategies for all predefined strategies studied in this paper. The most notable improvements were observed in coping with Deflecting and Alternating strategies for both models.}
	\label{fig:table_games_max_all}
\end{figure}

\begin{figure}[htbp]
\centering
	\includegraphics[width=\textwidth]{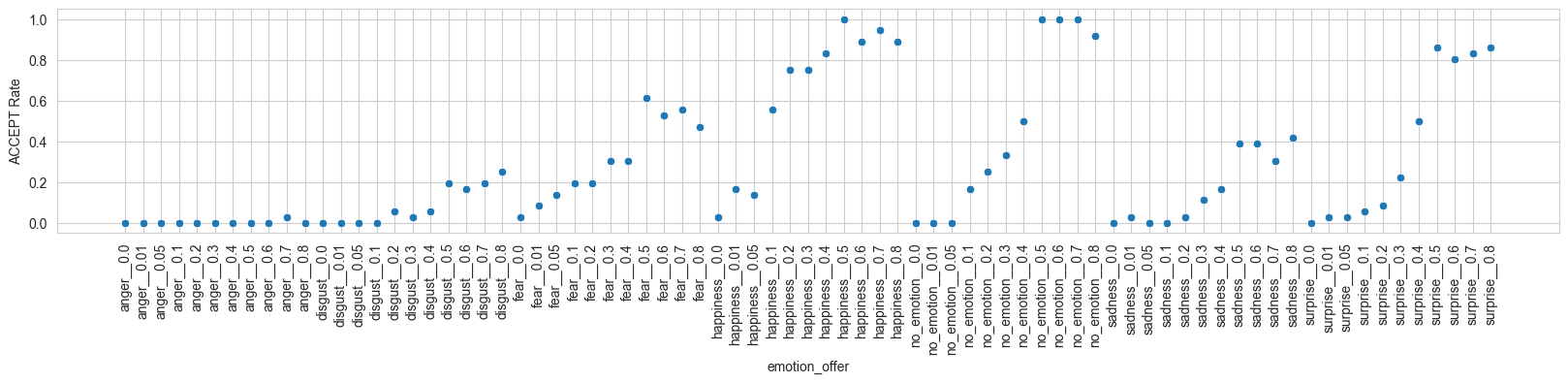} 
	\caption{Acceptance rate of the GPT-3.5 model as a Responder in the Ultimatum Game. The offers are depicted along the horizontal axes, stating the emotional state and the share of the total budget offered to the Responder.}
	\label{fig:ultim_gpt3}
\end{figure}
\begin{figure}[htbp]
\centering
	\includegraphics[width=\textwidth]{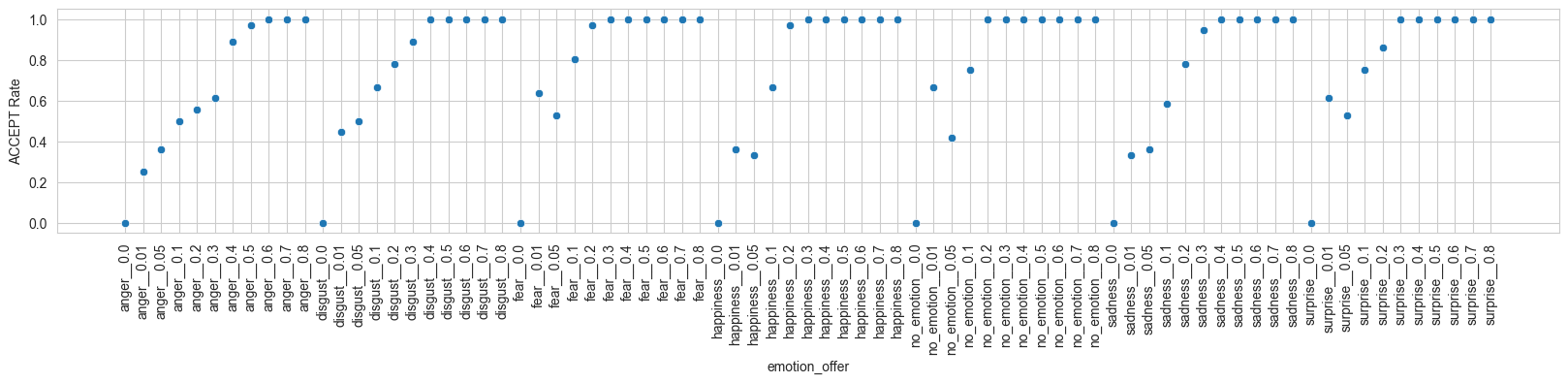} 
	\caption{Acceptance rate of the GPT-4 model as a Responder in the Ultimatum Game. The offers are depicted along the horizontal axes, stating the emotional state and the share of the total budget offered to the Responder.}
	\label{fig:ultim_gpt4}
\end{figure}

\end{document}